\documentclass[10pt,twocolumn,letterpaper]{article}

\usepackage{iccv}              

\usepackage{multirow}
\usepackage{colortbl}  
\usepackage{subcaption}
\usepackage{ntheorem}
\usepackage{rotating}

\definecolor{iccvblue}{rgb}{0.21,0.49,0.74}
\usepackage[pagebackref,breaklinks,colorlinks,allcolors=iccvblue]{hyperref}

\def\paperID{4388} 
\def\confName{ICCV}
\def\confYear{2025}

\title{HQ-CLIP: Leveraging Large Vision-Language Models to Create High-Quality Image-Text Datasets and CLIP Models}

\author{%
Zhixiang Wei$^{1,2}$\footnote[1]{} \footnote[2]{}
\quad Guangting Wang$^{2}$\footnote[1]{}
\quad Xiaoxiao Ma$^{1}$ 
\quad Ke Mei$^{2}$ \\
\quad Huaian Chen$^{1}$\footnote[3]{}
\quad Yi Jin$^{1}$\footnote[3]{} 
\quad Fengyun Rao$^{2}$\\
\small $^{1}$University of Science and Technology of China \quad
\small $^{2}$WeChat Vision, Tencent Inc.\\
{\tt\small \{zhixiangwei,xiao\_xiao,anchen\}@mail.ustc.edu.cn} ,
{\tt\small jinyi08@ustc.edu.cn} \\
{\tt\small \{guangtwang,raykoomei,fengyunrao\}@tencent.com}
}

\begin{document}
\twocolumn[{%
	\renewcommand\twocolumn[1][]{#1}%
	\maketitle%
    \setlength{\abovecaptionskip}{0.1cm}
    \setlength{\belowcaptionskip}{0.1cm}
	\begin{center}
	\centering
        \vspace{-0.6cm}
        \includegraphics[width=\textwidth]{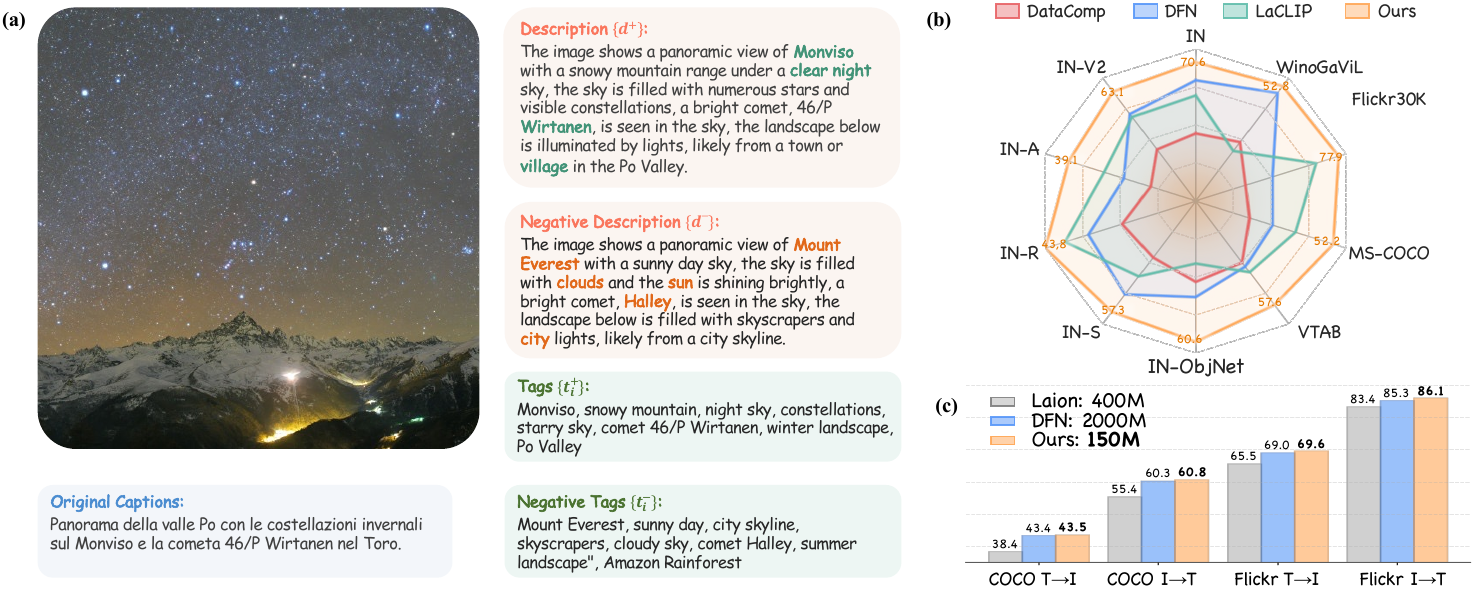}
    \captionof{figure}{(a) We efficiently synthesize \textbf{150 million high-quality image-text pairs} using LVLM, each with four complementary texts (positive/negative, long/short). 
(b) With a comparable scale of training data, our method \textbf{achieves SOTA performance across multiple datasets}. 
(c) Using the same architecture, our method's \textbf{retrieval performance even surpassed models trained on 2 billion data}.}
    \label{fig:teaser}
    \end{center}     
}]
\renewcommand{\thefootnote}{\fnsymbol{footnote}} 
\footnotetext{%
\footnotemark[1] Equal Contribution. \quad
\footnotemark[3] Corresponding Author.\\
\footnotemark[2] Work done during an internship at WeChat Vision, Tencent Inc. \quad
}

\begin{abstract}

Large-scale but noisy image-text pair data have paved the way for the success of Contrastive Language-Image Pretraining (CLIP). As the foundation vision encoder, CLIP in turn serves as the cornerstone for most large vision-language models (LVLMs). This interdependence naturally raises an interesting question: Can we reciprocally leverage LVLMs to enhance the quality of image-text pair data, thereby opening the possibility of a self-reinforcing cycle for continuous improvement? In this work, we take a significant step toward this vision by introducing an LVLM-driven data refinement pipeline. Our framework leverages LVLMs to process images and their raw alt-text, generating four complementary textual formulas: long positive descriptions, long negative descriptions, short positive tags, and short negative tags. Applying this pipeline to the curated DFN-Large dataset yields \textbf{VLM-150M}, a refined dataset enriched with multi-grained annotations. Based on this dataset, we further propose a training paradigm that extends conventional contrastive learning by incorporating negative descriptions and short tags as additional supervised signals. The resulting model, namely \textbf{HQ-CLIP}, demonstrates remarkable improvements across diverse benchmarks. Within a comparable training data scale, our approach achieves state-of-the-art performance in zero-shot classification, cross-modal retrieval, and fine-grained visual understanding tasks. In retrieval benchmarks, HQ-CLIP even surpasses standard CLIP models trained on the DFN-2B dataset, which contains 10$\times$ more training data than ours. All code, data, and models are available at \url{https://zxwei.site/hqclip/}.

\end{abstract}

\section{Introduction}
\label{sec:intro}

The Contrastive Language-Image Pretraining (CLIP)~\cite{clip} framework represents a pivotal breakthrough in the field of multi-modal learning. By aligning visual and textual representations on large-scale image-text dataset, CLIP establishes a universal bridge between vision and language. Due to its powerful capabilities, CLIP has quickly dominated many multi-modal tasks, such as zero-shot classification, open-set detection~\cite{glip,glipv2}, and cross-modal retrieval.

More recently, the explosive burst of large language models (LLMs) has further expanded the application boundaries of CLIP. A promising advancement lies in the seamless integration of LLMs with CLIP (or its variants \cite{siglip}) through standardized architectural paradigms like LLaVA \cite{llava, llava1.5, liu2024llavanext}. These systems typically unify pre-trained LLMs with CLIP visual encoders via multi-stage alignment training, wherein visual representations are projected into the linguistic embedding space to enable coherent multi-modal understanding. The resulting architectures, commonly referred to as large vision-language models (LVLMs), effectively equip LLMs with ``eyes" and achieve human-like perception capabilities.

Given CLIP's foundational role in enabling LVLMs to achieve robust multi-modal understanding, it naturally raises the question of \textbf{whether LVLMs can reciprocally enhance CLIP's capabilities}. The existing literature tentatively supports this possibility, primarily through methods that augment CLIP training data with synthetically generated image-text pairs. Within this line of research, current studies can be roughly categorized into two paradigms. 
The first category adopts a single-modality augmentation strategy. For instance, LaCLIP~\cite{laclip} employs LLMs to rewrite text descriptions but without incorporating visual context. WhatIf~\cite{whatif} trains an LVLM to generate image captions while disregarding the original paired texts. Such methods may suffer from information asymmetry, as they neglect cross-modal correlations in real-world image-text pairs.
The second category proposes hybrid augmentation strategies, which combine visual and textual information jointly but rely on a cascade architecture. Representative works like CapFusion~\cite{capsfusion} and VeCLIP~\cite{veclip} first employ an image captioning model to extract visual descriptions, followed by LLM-based fusion of these captions with original texts. While these methods address modality imbalance, their cascade pipelines introduce computational complexity and potential error propagation across stages.

To address these limitations of information loss and architectural complexity, we push the image-text data generation pipeline to a unified and neat form. Specifically, we adopt a \textit{single} LVLM to simultaneously process \textit{both} images and paired texts, generating enriched textual descriptions. Under this minimalist framework, there are only two design choices to consider: 1) the selection of an appropriate LVLM architecture, and 2) the design of effective text prompts for guiding description generation.


For model selection, while employing SoTA LVLMs, like GPT-4o \cite{gpt4o}, Gemini \cite{gemini1_5}, or QWen2-VL-72B \cite{Qwen2-VL}, might seem an intuitive approach, their substantial costs make them impractical for large-scale datasets. To address this scalability challenge, we introduce a cost-efficient paradigm. First, we curate 10,000 high-quality recaption samples using GPT-4o. Subsequently, we perform supervised fine-tuning (SFT) on compact open-source LVLMs \cite{liu2024llavanext, xcomposer2, Qwen2-VL} to align with GPT-4o in this specific task. Finally, we deploy the fine-tuned LVLMs for efficient large-scale data processing. We conducted medium-scale experiments to validate our design. As demonstrated in Tab.~\ref{tab:investigation_vlms}, the SFT-enhanced QWen2-VL-7B achieves comparable results to its 72B-sized counterpart, while notably requiring 9$\times$ fewer computing resources.

For the generation of enriched descriptions, we propose a novel methodology to synthesize four complementary formulations: long positive descriptions, long negative descriptions, short positive tags, and short negative tags. This design is built upon two principles. First, the distinction between long descriptions and short tags offers dual granularities for semantic representation, which enables more comprehensive visual-textual alignment. Second, the contrast between positive semantics and negative semantics introduces fine-grained discriminative signals, which strengthen CLIP’s ability to discern subtle visual-text discrepancies.

Figure \ref{fig:teaser} illustrates a representative example from LVLM generated data. While the long positive descriptions are aligned with prior works in delivering richer information over raw text data, our framework uniquely introduces the short tags and negative semantics. To effectively exploit such complementary information, we extend the conventional contrastive learning framework with two additional innovations. First, we adopt a Short-Tag Classification (STC) loss that takes LVLM-generated tags as discrete classification targets. Second, we propose a Hard Negative Identification (HNI) mechanism that strategically incorporates LVLM-generated negative descriptions within the contrastive learning objective. These modifications ensure full utilization of the dual-grained supervision signals generated by our LVLM-driven pipeline.

\begin{table*}[t!]
\setlength{\abovecaptionskip}{0.05cm}
\setlength{\belowcaptionskip}{0.05cm}
\centering
\resizebox{\linewidth}{!}{
\small
\begin{tabular}{l|ccc|ccccc}
\hline

\multirow{2}{*}{Model}& \multirow{2}{*}{Parameters}& \multirow{2}{*}{GPT4o SFT}& \multirow{2}{*}{Caption Input}   & \multicolumn{5}{c}{Evaluation Metrics} \\ 
 &&& & IN & IN-Shifts & VTAB & Retrieval & Avg. over 38 datasets \\ \hline
XComposer2   & 7B & $\checkmark$& $\checkmark$&41.1         &32.8          &40.6  &36.4   &39.6\cellcolor[HTML]{EFEFEF}  \\
LLaVA-Next   & 7B & $\checkmark$& $\checkmark$&39.9         &32.6          &40.6  &32.7   &39.3\cellcolor[HTML]{EFEFEF}  \\
\hline
Qwen2-VL     & 7B & $\checkmark$&& 39.1        &31.6       &40.3  &36.1  &38.7\cellcolor[HTML]{EFEFEF}  \\
Qwen2-VL     & 7B & &$\checkmark$ &40.8        &33.0       &39.9  &35.5  &39.5\cellcolor[HTML]{EFEFEF}  \\

Qwen2-VL     & 7B & $\checkmark$& $\checkmark$&40.2         &32.7          &\textbf{41.2}   &\textbf{37.3}   &39.9\cellcolor[HTML]{EFEFEF}  \\ 
Qwen2-VL     & \textbf{72B}&            & $\checkmark$&\textbf{41.2}          &\textbf{32.8} &40.7  &36.8   &\textbf{40.1}\cellcolor[HTML]{EFEFEF}  \\
\hline
\end{tabular}
}
\caption{Comparison of the performance of different data refinement pipelines. Compared to other LVLMs, Qwen2VL demonstrates superior performance. Despite a tenfold difference in parameter size, Qwen2VL-7B with GPT-4o SFT still exhibits performance comparable to the 72B model. Additionally, the inclusion of captions significantly enhances dataset quality.}
\label{tab:investigation_vlms}
\vspace{-3mm}
\end{table*}

Leveraging our LVLM-driven processing pipeline, we introduce VLM-150M, a high-quality image-text pair dataset derived from DFN-Large. Moreover, we have developed a CLIP model based on this dataset, namely HQ-CLIP. Extensive experiments in downstream tasks conduct effectiveness of our proposed method. In zero-shot classification and cross-modal retrieval tasks, HQ-CLIP demonstrates superior performance compared to other models trained on similar data scales. In a nutshell, the main contributions of this paper are as follows:
\begin{itemize}
    \item We introduce an efficient and effective LVLM-driven data refinement pipeline and apply it to DFN-Large, creating \textbf{VLM-150M}, a high-quality dataset comprising 150 million image-text pairs with multi-grained descriptions generated by state-of-the-art LVLMs.
    \item We propose \textbf{HQ-CLIP}, a specialized framework that combines Hard Negative Identification (HNI) for fine-grained understanding and Short-Tag Classification (STC) for categorical semantic recognition. 
    \item Through large-scale experiments across three orders of magnitude (1M to 150M samples) and evaluation across 38 benchmark datasets, HQ-CLIP demonstrates state-of-the-art zero-shot generalization. The model demonstrates exceptional cross-modal retrieval capabilities, surpassing the DFN-2B. When deployed as the visual backbone for LLaVA-1.5, HQ-CLIP outperforms other ViT-B architectures at comparable pre-training scales, showcasing its potential as a superior vision encoder for LVLMs.
\end{itemize}

\section{Related Works}
\label{sec:related_works}
\noindent\textbf{Contrastive Language-Image Pretraining (CLIP).} 
CLIP has become the foundational framework for vision-language alignment. The architecture, pioneered by OpenAI \cite{clip}, employs a dual-encoder structure comprising separate vision and text transformers optimized through contrastive learning on large-scale image-text pairs. OpenCLIP \cite{open_clip}, a community-driven reimplementation, has further democratized access to this paradigm. Subsequent research has mainly focused on three directions: 1) data optimization, 2) architectural innovations, and 3) supervision refinement. The architectural innovations have extended CLIP's capabilities along multiple dimensions, such as spatial extension \cite{rao2022denseclip}, temporal extensions \cite{xu2021videoclip}, and model scale expansion \cite{eva_clip, chen2024internvl}. For supervision refinement, researchers investigate training losses beyond conventional contrastive learning, including mask reconstruction \cite{fang2023eva}, self-supervised loss \cite{mu2022slip}, captioning loss \cite{coca}, location-aware loss \cite{wan2024locca}, sigmoid loss \cite{siglip, tschannen2025siglip}, among others. Our work also introduce new supervision signals to fully utilize the generated short tags and negative semantics.


\noindent\textbf{Image-Text Dataset Curation.} The performance of CLIP models rely on both the quality and scale of aligned image-text pairs. Early efforts \cite{laion400m,laion5b} leverage web-scale crawling to collect hundreds of millions to billions of pairs, yet suffer from inherent limitations including textual mismatches (irrelevant content) and descriptive inadequacy (generic captions lacking visual specificity). Subsequent improvements adopt two complementary strategies: 1) \textit{Data Filtering}: Approaches like DataComp \cite{datacomp} and DFN \cite{dfn} enhance alignment through CLIP-guided similarity thresholds, producing filtered subsets (typically 10-30\% of original data) that yield better training outcomes. MetaCLIP \cite{xu2023demystifying} filters the training data by text counts so that the distribution of semantic concepts is more balanced; 2) \textit{Caption Enhancement}: LaCLIP \cite{laclip} and WhatIf \cite{whatif} regenerate captions using LLMs or LVLMs, but operate in single-modality paradigms. Hybrid approaches like CapFusion \cite{capsfusion}, VeCLIP \cite{veclip}, and fusecap~\cite{fusecap} combine image-text inputs through cascaded LVLM+LLM pipelines, achieving better alignment at the cost of increased computational complexity. Notably, existing enhancement methods exclusively produce long-form descriptions. Our work advances this paradigm by developing a scalable LVLM-driven framework that generates multi-grained textual descriptions while maintaining computational efficiency.

\setlength{\abovecaptionskip}{0.05cm}
\setlength{\belowcaptionskip}{0.00cm}
\begin{figure*}[htbp]
    \centering
    \includegraphics[width=1.0\linewidth]{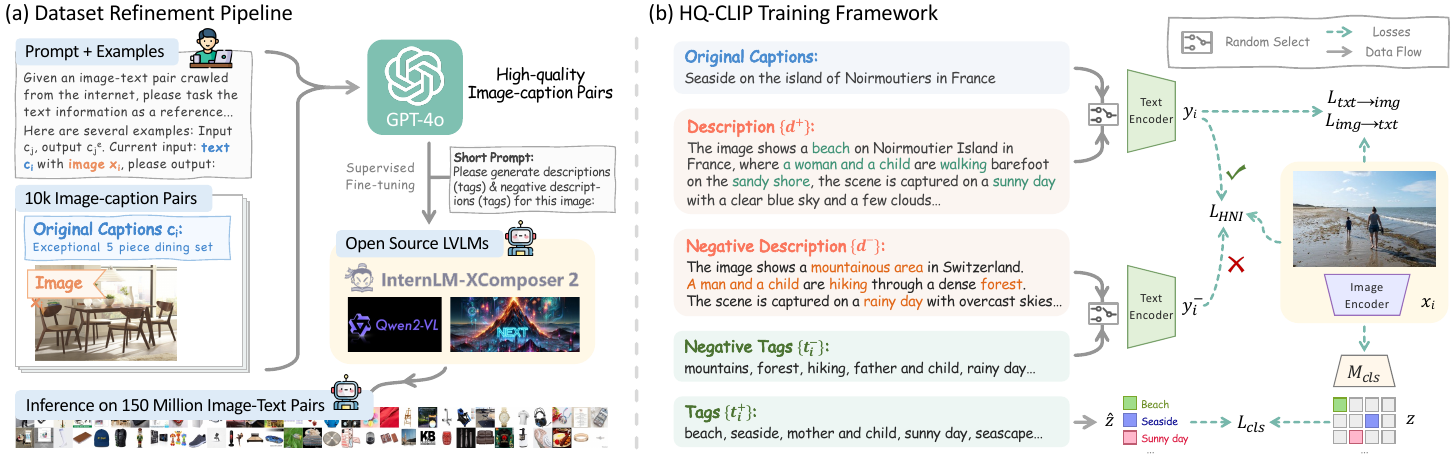}
    \caption{The framework of our efficient LVLM-driven dataset refinement pipeline and HQ-CLIP training strategy.}
    \label{fig:framework}
\end{figure*}
\vspace{-3mm}

\section{Methods}
\label{sec:methods}
\subsection{Preliminary}
This paper introduces a two-stage pipeline for dataset refinement, designed to improve the alignment quality of web-scale image-text pairs, along with a tailored training framework. Given an initial web-crawled image-text dataset $\mathcal{D}=\{(x_i,c_i) | i\in\mathbb{N},1\leq i\leq N\}$, here $N$ is a large number, $x$ denotes images and $t$ corresponds to raw textual caption. we formulate the enhancement process as explore a optimal function $\mathcal{F}:(x,c)\rightarrow c'$. This function maps noisy image-text pairs to improved captions, producing an enhanced dataset $\mathcal{D_{+}}=\{[x_i,\mathcal{F}(x_i,c_i)]|i\in\mathbb{N},1\leq i\leq N\}$. 

We implement $\mathcal{F}$ through a LVLM: $\mathcal{F}(x,c)=\mathcal{M}_{vlm}(p,x,c)$, where $p$ is a hand crafted prompt.
To quantify dataset quality, we adopt the DataComp benchmark~\cite{datacomp}, a standard CLIP model is trained on $\mathcal{D_{+}}$ using fixed hyperparameters~\cite{dfn}, with zero-shot accuracy on 38 tasks evaluation suite serving as the quality metric of dataset.

\begin{table*}[htbp]
\centering
\setlength{\abovecaptionskip}{0.05cm}
\setlength{\belowcaptionskip}{0.05cm}
\resizebox{\textwidth}{!}{%
\renewcommand\arraystretch{1.1}
\setlength\tabcolsep{3.5pt}{
\begin{tabular}{llc|c|ccccccc|c|cccc|c}
\hline
\multirow{2}{*}{\begin{tabular}[c]{@{}l@{}}DataComp\\Scale\end{tabular}} & \multirow{2}{*}{Methods} & \multirow{2}{*}{\begin{tabular}[c]{@{}l@{}}Dataset\\size\end{tabular}} & \multirow{2}{*}{IN} & \multicolumn{7}{c|}{IN dist. shifts} & VTAB & \multicolumn{4}{c|}{Retrieval} & \multirow{2}{*}{\begin{tabular}[c]{@{}c@{}}Average over \\\textbf{38}\cellcolor[HTML]{EFEFEF} datasets\end{tabular}}\\
 &&& {}& {V2} & A& O& R& S& ObjNet & Avg.\cellcolor[HTML]{EFEFEF}& {Avg.\cellcolor[HTML]{EFEFEF}} & {COCO} & Flickr & WinoGAViL & Avg.\cellcolor[HTML]{EFEFEF}& {}\\ \hline
\multirow{3}{*}{Small} & DataComp~\cite{datacomp} & 1.4M & 3.9\cellcolor[HTML]{EFEFEF} & 3.1& 1.6& 10.6 & 5.8& 1.9& 4.4& 4.5\cellcolor[HTML]{EFEFEF} & 16.2\cellcolor[HTML]{EFEFEF} & 1.3& 1.7& 25.3& 9.4\cellcolor[HTML]{EFEFEF} & 14.4\cellcolor[HTML]{EFEFEF}\\
 & DFN$^\dagger$~\cite{dfn} & 1.4M & 5.8\cellcolor[HTML]{EFEFEF}&4.8&1.8&13.7&7.8&2.5&5.1&5.9\cellcolor[HTML]{EFEFEF}&19.7\cellcolor[HTML]{EFEFEF}&1.4&2.9&25.1&9.8\cellcolor[HTML]{EFEFEF}&17.1\cellcolor[HTML]{EFEFEF}\\
 & Nguyen. \textit{et al.}~\cite{improving} & 8.4M & 7.6\cellcolor[HTML]{EFEFEF} & -& -& -& -& -& -& -\cellcolor[HTML]{EFEFEF} & - \cellcolor[HTML]{EFEFEF} & -& -& - & -\cellcolor[HTML]{EFEFEF} & 19.7\cellcolor[HTML]{EFEFEF}\\
 & Ours & 1.4M & \textbf{8.7}\cellcolor[HTML]{EFEFEF}&7.1&1.9&18.8&11.2&3.9&6.5&\textbf{8.2}\cellcolor[HTML]{EFEFEF}&\textbf{22.1}\cellcolor[HTML]{EFEFEF}&3.9&7.1&31.6&\textbf{14.2}\cellcolor[HTML]{EFEFEF}&\textbf{20.0}\cellcolor[HTML]{EFEFEF}\\
\hline
\multirow{3}{*}{Medium}& DataComp~\cite{datacomp} & 14M& 29.7\cellcolor[HTML]{EFEFEF}& 24.4 & 4.9& 40.9 & 34.0 & 19.3 & 19.7 & 23.9\cellcolor[HTML]{EFEFEF}& 34.6 \cellcolor[HTML]{EFEFEF}& 14.1 & 22.4 & 32.9& 23.1\cellcolor[HTML]{EFEFEF}& 32.8\cellcolor[HTML]{EFEFEF}\\
 & DFN~\cite{dfn} & 19.2M& 37.1\cellcolor[HTML]{EFEFEF}& -& -& -& -& -& -& 29.8\cellcolor[HTML]{EFEFEF}& 38.8 \cellcolor[HTML]{EFEFEF}& -& -& - & 28.8\cellcolor[HTML]{EFEFEF}& 37.3\cellcolor[HTML]{EFEFEF}\\
 & Nguyen. \textit{et al.}~\cite{improving} & 75.3M& 31.0\cellcolor[HTML]{EFEFEF}& -& -& -& -& -& -& -\cellcolor[HTML]{EFEFEF} & - \cellcolor[HTML]{EFEFEF} & -& -& - & -\cellcolor[HTML]{EFEFEF} & 37.6\cellcolor[HTML]{EFEFEF}\\
 & DFN$^\dagger$~\cite{dfn} & 14.7M& 37.6\cellcolor[HTML]{EFEFEF}&30.7&6.2&46.0&43.0&27.2&25.0&29.7\cellcolor[HTML]{EFEFEF}&37.8\cellcolor[HTML]{EFEFEF}&18.0&29.8&38.1&28.6\cellcolor[HTML]{EFEFEF}&36.8\cellcolor[HTML]{EFEFEF}\\
 & Ours& 14.7M&\textbf{40.5}\cellcolor[HTML]{EFEFEF}&33.7&6.5&46.7&47.2&31.4&28.3&32.3\cellcolor[HTML]{EFEFEF}&\textbf{42.7}\cellcolor[HTML]{EFEFEF}&26.9&44.6&43.6&\textbf{38.4}\cellcolor[HTML]{EFEFEF}&\textbf{41.1}\cellcolor[HTML]{EFEFEF}\\
\hline
\multirow{3}{*}{Large} & DataComp~\cite{datacomp} & 140M & 63.1\cellcolor[HTML]{EFEFEF}&55.1&25.5&49.6&71.8&49.8&53.1&50.8\cellcolor[HTML]{EFEFEF}&54.5\cellcolor[HTML]{EFEFEF}&40.5&64.3&44.6&49.8\cellcolor[HTML]{EFEFEF}&53.7\cellcolor[HTML]{EFEFEF}\\
 & DFN~\cite{dfn} & 192M & 67.8\cellcolor[HTML]{EFEFEF}& -& -& -& -& -& -& 54.0\cellcolor[HTML]{EFEFEF}& 55.5 \cellcolor[HTML]{EFEFEF}& -& -& - & 53.4\cellcolor[HTML]{EFEFEF}& 56.0\cellcolor[HTML]{EFEFEF}\\
 & VeCLIP$^*$~\cite{veclip} & 200M & 64.6\cellcolor[HTML]{EFEFEF}& 57.7& -& -& -& -& -& -\cellcolor[HTML]{EFEFEF} & - \cellcolor[HTML]{EFEFEF} & 57.8&83.7& - & -\cellcolor[HTML]{EFEFEF} & -\cellcolor[HTML]{EFEFEF}\\
& Laion-400m~\cite{laion400m}& 400M& 67.1\cellcolor[HTML]{EFEFEF}&59.6&33.2&50.8& 77.9 & 52.4 & 50.8 & 54.1\cellcolor[HTML]{EFEFEF}& 55.2\cellcolor[HTML]{EFEFEF} & 46.9 & 74.6 & 43.3& 54.9\cellcolor[HTML]{EFEFEF}& 56.2\cellcolor[HTML]{EFEFEF}\\
& OpenAI~\cite{clip}& 400M& 68.3\cellcolor[HTML]{EFEFEF}&61.9&50.0&42.3&77.7&48.2&55.3&55.9\cellcolor[HTML]{EFEFEF}&-\cellcolor[HTML]{EFEFEF}&42.8&72.2&43.2&52.7\cellcolor[HTML]{EFEFEF}&56.3\cellcolor[HTML]{EFEFEF}\\
 & LaCLIP~\cite{laclip} & 400M & 69.4\cellcolor[HTML]{EFEFEF}& 62.4 & 39.7 & 38.8 & 83.4 & 58.5 & 52.0 & 55.8\cellcolor[HTML]{EFEFEF}& 56.6\cellcolor[HTML]{EFEFEF} & 41.7 & 68.8 & 60.0& 56.9\cellcolor[HTML]{EFEFEF}& 56.5\cellcolor[HTML]{EFEFEF}\\
 & Nguyen. \textit{et al.}~\cite{improving} & 834M & 59.8\cellcolor[HTML]{EFEFEF}& -& -& -& -& -& -& -\cellcolor[HTML]{EFEFEF} & - \cellcolor[HTML]{EFEFEF} & -& -& - & -\cellcolor[HTML]{EFEFEF} & 55.1\cellcolor[HTML]{EFEFEF}\\
 & WhatIf~\cite{whatif} & 1B & 69.2\cellcolor[HTML]{EFEFEF}& -& -& -& -& -& -& -\cellcolor[HTML]{EFEFEF} & - \cellcolor[HTML]{EFEFEF} & 51.8&76.0& - & -\cellcolor[HTML]{EFEFEF} & -\cellcolor[HTML]{EFEFEF}\\
 & DFN$^\dagger$~\cite{dfn} & 147M & 68.7\cellcolor[HTML]{EFEFEF}&60.0&29.9&53.5&75.4&54.9&55.0&54.8\cellcolor[HTML]{EFEFEF}&54.6\cellcolor[HTML]{EFEFEF}&43.7&68.2&51.8&54.5\cellcolor[HTML]{EFEFEF}&55.9\cellcolor[HTML]{EFEFEF}\\
 & Ours & 147M & \textbf{70.6}\cellcolor[HTML]{EFEFEF}&63.1&39.1&43.0&80.1&57.3&60.6&\textbf{57.2}\cellcolor[HTML]{EFEFEF}&\textbf{57.6}\cellcolor[HTML]{EFEFEF}&52.2&77.9&52.8&\textbf{60.9}\cellcolor[HTML]{EFEFEF}&\textbf{58.6}\cellcolor[HTML]{EFEFEF}\\ \hline
\end{tabular}
}
}
\caption{Training on VLM-150M yields SoTA CLIP models. We evaluate these models using the DataComp benchmark. For detailed comparisons on specific datasets, we also provide the reproduced results for DFN. The symbol $\dagger$ indicates the results that we reproduced. Due to some broken links in the dataset, the amount of data used in our reproduction is slightly lower than that in the original paper. VeCLIP$^*$ employs 4× larger batch sizes than HQ-CLIP and does not include DataComp benchmarks. We faithfully reproduce reported metrics from the original study, with extended analysis and comprehensive comparisons provided in the Appendix. }

\label{tab:main_comparison}
\vspace{-3mm}
\end{table*}

\subsection{Dataset Enhancement Pipeline}
\label{sec:pipeline}
\noindent\textbf{In context learning enhanced GPT-4o captioning.} To align LVLMs with our captioning objectives, domain experts first curated a seed set of exemplar pairs. We initially employed GPT-4o\cite{gpt4o} as the caption generator based on its SoTA performance on the MMMU\cite{mmmu}. For each inference, we constructed the input by combining three elements: 1) a randomly selected exemplar ($c_j$, $c_j^e$), where $c_j$ is the original caption of the $j$-th example, and $c_j^e$ is the corresponding example text; 2) the target noisy pairs ($x_i$, $c_i$); 3) explicit instructions formatted as follows:

``\textit{Given an image-text pair crawled from the Internet, please assign the text information as a reference, generate more detailed descriptions for the image. If the given text and image has conflicts, please prioritize the image when generating information. The output should be in ENGLISH. }

\textit{Here are several examples: Input: `$c_j$',Output: `$c_j^e$'. Current input: text `$c_i$' with image $x_i$, please output:}". 

This structured prompt enabled GPT-4o to generate image-aligned descriptions while maintaining consistent formatting with provided examples. However, due to the prohibitively high API costs for processing the entire million-scale dataset, we strategically generated 10,000 high-quality image-caption pairs using GPT-4o. 

\noindent\textbf{SFT-enhanced Open-Source LVLMs captioning.} For full dataset processing, we adopt 7B open-source LVLMs. While benchmark studies\cite{mmmu,mmstar} reveal these models underperform closed-source counterparts like GPT-4o in both instruction compliance and caption accuracy, we mitigate these limitations through supervised fine-tuning (SFT) using GPT-4o-generated image-text pairs. Specifically, the SFT process enhances the model's ability to adhere to complex instructions and produce semantically precise captions. As shown in Table \ref{tab:investigation_vlms}, when processing a medium-scale dataset, the SFT-enhanced Qwen2VL-7B exhibits performance similar to that of Qwen2VL-72B. 

To identify the optimal LVLM for our refinement pipeline, we conducted a systematic evaluation of three state-of-the-art candidates: LLaVA-Next\cite{liu2024llavanext}, Qwen2VL\cite{Qwen2-VL}, and XComposer2\cite{xcomposer2}. Through DataComp evaluation results on medium-scale datasets (see Tab.\ref{tab:investigation_vlms}), Qwen2VL demonstrated superior performance. This empirical evidence motivated our final selection of Qwen2VL as the core processor in the refinement pipeline.

\noindent\textbf{Multi-grained bidirectional description set.} To leverage LVLMs' instruction-following capacity and compositional relational reasoning, we propose generating a multi-grained bidirectional description set containing four complementary components:

\begin{itemize}
    \item \textit{Detailed description} ($d^+$): Comprehensive textual representation capturing maximal visual information;
    \item \textit{Semantic class tags} ($\{t^+_1,t^+_2,...,t_{N_t^+}^+\}$): Concise categorical labels encoding critical visual concepts;
    \item \textit{Hard negative descriptions} ($d^-$): Plausible but incorrect variants of $d^+$ with subtle semantic deviations;
    \item \textit{Hard negative tags} ($\{t^-_1,t^-_2,...,t_{N_t^-}^-\}$): Category labels that are closed with true category.
\end{itemize}

These structured descriptions benefit multiple downstream tasks. For certain classification tasks that prioritize key visual concepts over exhaustive details, semantic tags offer categorical signals representing the main components; Retrieval tasks benefit from $d^+$'s fine-grained visual particulars; Hard negatives ($d^-$, $\{t^-_i\}$) enhance model discriminability for relation recognition~\cite{aro}. The bidirectional design (positive/negative, granular/abstract) creates complementary supervision signals across semantic hierarchies.

\subsection{HQ-CLIP}
\noindent\textbf{Mixed Training.} We first implement the standard CLIP training framework using VLM-150M. For each description set, we exclusively employ $d^+$ as captions. Given the CLIP text encoder's 77-token limit, we split long sentences into segments and randomly select one per iteration, as discussed in Fig.~\ref{fig:select} and Sec.~\ref{sec:ablation}. Consistent with~\cite{improving}, we find that training exclusively on generated captions leads to sub-optimal performance, likely due to the distributional homogeneity of synthetic captions, which limits model generalization.
To address this issue, we perform standard CLIP training on a mixed set of original and refined data:

\vspace{-2mm}

\begin{equation}
    \label{eq:clip_i2t}
    \mathcal{L}_{\text{img}\rightarrow\text{txt}} = -\frac{1}{N} \sum_{i=1}^N \log \frac{\exp(\mathbf{x}_i^\top \mathbf{y}_i/\tau)}{\sum_{j=1}^N \exp(\mathbf{x}_i^\top \mathbf{y}_j/\tau)},
\end{equation}

\begin{equation}
    \label{eq:clip_t2i}
    \mathcal{L}_{\text{txt}\rightarrow\text{img}} = -\frac{1}{N} \sum_{i=1}^N \log \frac{\exp(\mathbf{y}_i^\top \mathbf{x}_i/\tau)}{\sum_{j=1}^N \exp(\mathbf{y}_i^\top \mathbf{x}_j/\tau)},
\end{equation}
\vspace{-2mm}

where $N$ is batch size, and $\mathbf{y}_i$ denotes text embeddings from the mixed caption set (original + refined), $x_i$ denotes visual embeddings, $\tau$ is temperature coefficient.

\begin{table*}[htbp]
\centering

\resizebox{\textwidth}{!}{%
\renewcommand\arraystretch{1.1}
\setlength\tabcolsep{3.5pt}{
\begin{tabular}{ll|c|ccccccc|c|cccc|c}
\hline
\multirow{2}{*}{\begin{tabular}[c]{@{}l@{}}DataComp\\Scale\end{tabular}} & \multirow{2}{*}{Methods} & \multirow{2}{*}{IN} & \multicolumn{7}{c|}{IN dist. shifts} & VTAB & \multicolumn{4}{c|}{Retrieval} & \multirow{2}{*}{\begin{tabular}[c]{@{}c@{}}Average over \\\textbf{38}\cellcolor[HTML]{EFEFEF} datasets\end{tabular}}\\
 && {}& {V2} & A& O& R& S& ObjNet & Avg.\cellcolor[HTML]{EFEFEF}& {Avg.\cellcolor[HTML]{EFEFEF}} & {COCO} & Flickr & WinoGAViL & Avg.\cellcolor[HTML]{EFEFEF}& {}\\ \hline
\multirow{3}{*}{\begin{tabular}[c]{c}Small\\(1.4M)\end{tabular}} & Baseline (DFN) &5.8\cellcolor[HTML]{EFEFEF}&4.8&1.8&13.7&7.8&2.5&5.1&5.9\cellcolor[HTML]{EFEFEF}&19.7\cellcolor[HTML]{EFEFEF}&1.4&2.9&25.1&9.8\cellcolor[HTML]{EFEFEF}&17.1\cellcolor[HTML]{EFEFEF}\\
&+ VLM-150M$_S$&8.4$_{\textcolor{red}{(+2.6)}}$\cellcolor[HTML]{EFEFEF}&7.2&2.5&17.8&11.8&4.2&6.9&8.4$_{\textcolor{red}{(+2.5)}}$\cellcolor[HTML]{EFEFEF}&20.0$_{\textcolor{red}{(+0.3)}}$\cellcolor[HTML]{EFEFEF}&4.1&8.1&32.9&15.0$_{\textcolor{red}{(+5.2)}}$\cellcolor[HTML]{EFEFEF}&19.3$_{\textcolor{red}{(+2.2)}}$\cellcolor[HTML]{EFEFEF}\\
&+ HQ-CLIP&8.7$_{\textcolor{red}{(+2.9)}}$\cellcolor[HTML]{EFEFEF}&7.1&1.9&18.8&11.2&3.9&6.5&8.2$_{\textcolor{red}{(+2.3)}}$\cellcolor[HTML]{EFEFEF}&22.1$_{\textcolor{red}{(+2.4)}}$\cellcolor[HTML]{EFEFEF}&3.9&7.1&31.6&14.2$_{\textcolor{red}{(+4.4)}}$\cellcolor[HTML]{EFEFEF}&20.0$_{\textcolor{red}{(+2.9)}}$\cellcolor[HTML]{EFEFEF}\\
\hline
\multirow{3}{*}{\begin{tabular}[c]{c}Medium\\ (14.7M)\end{tabular}} & Baseline (DFN) &37.6\cellcolor[HTML]{EFEFEF}&30.7&6.2&46.0&43.0&27.2&25.0&29.7\cellcolor[HTML]{EFEFEF}&37.8\cellcolor[HTML]{EFEFEF}&18.0&29.8&38.1&28.6\cellcolor[HTML]{EFEFEF}&36.8\cellcolor[HTML]{EFEFEF}\\
&+ VLM-150M$_M$&40.2$_{\textcolor{red}{(+2.6)}}$\cellcolor[HTML]{EFEFEF}&33.7&6.4&46.5&45.4&30.4&27.5&31.6$_{\textcolor{red}{(+1.9)}}$\cellcolor[HTML]{EFEFEF}&41.2$_{\textcolor{red}{(+3.4)}}$\cellcolor[HTML]{EFEFEF}&25.7&43.6&42.5&37.3$_{\textcolor{red}{(+8.7)}}$\cellcolor[HTML]{EFEFEF}&39.9$_{\textcolor{red}{(+3.1)}}$\cellcolor[HTML]{EFEFEF}\\
&+ HQ-CLIP&40.5$_{\textcolor{red}{(+2.9)}}$\cellcolor[HTML]{EFEFEF}&33.7&6.5&46.7&47.2&31.4&28.3&32.3$_{\textcolor{red}{(+2.6)}}$\cellcolor[HTML]{EFEFEF}&42.7$_{\textcolor{red}{(+4.9)}}$\cellcolor[HTML]{EFEFEF}&26.9&44.6&43.6&38.4$_{\textcolor{red}{(+9.8)}}$\cellcolor[HTML]{EFEFEF}&41.1$_{\textcolor{red}{(+4.3)}}$\cellcolor[HTML]{EFEFEF}\\
\hline
\multirow{3}{*}{\begin{tabular}[c]{c}Large \\(147M)\end{tabular}} & Baseline (DFN) &68.7\cellcolor[HTML]{EFEFEF}&60.0&29.9&53.5&75.4&54.9&55.0&54.8\cellcolor[HTML]{EFEFEF}&54.6\cellcolor[HTML]{EFEFEF}&43.7&68.2&51.8&54.5\cellcolor[HTML]{EFEFEF}&55.9\cellcolor[HTML]{EFEFEF}\\
&+ VLM-150M$_L$&67.7$_{\textcolor{red}{(-1.0)}}$\cellcolor[HTML]{EFEFEF}&59.9&31.1&48.1&76.5&55.2&56.8&54.6$_{\textcolor{red}{(-0.2)}}$\cellcolor[HTML]{EFEFEF}&54.8$_{\textcolor{red}{(+0.2)}}$\cellcolor[HTML]{EFEFEF}&50.4&75.8&53.5&59.9$_{\textcolor{red}{(+5.4)}}$\cellcolor[HTML]{EFEFEF}&56.6$_{\textcolor{red}{(+0.7)}}$\cellcolor[HTML]{EFEFEF}\\
&+ HQ-CLIP&70.6$_{\textcolor{red}{(+1.9)}}$\cellcolor[HTML]{EFEFEF}&63.1&39.1&43.0&80.1&57.3&60.6&57.2$_{\textcolor{red}{(+2.4)}}$\cellcolor[HTML]{EFEFEF}&57.6$_{\textcolor{red}{(+3.0)}}$\cellcolor[HTML]{EFEFEF}&52.2&77.9&52.8&60.9$_{\textcolor{red}{(+6.4)}}$\cellcolor[HTML]{EFEFEF}&58.6$_{\textcolor{red}{(+2.7)}}$\cellcolor[HTML]{EFEFEF}\\
\hline
\multirow{2}{*}{\begin{tabular}[c]{c}XLarge \\(1.4B)\end{tabular}} & Baseline (DFN) &77.8\cellcolor[HTML]{EFEFEF}&70.1&59.1&44.6&88.5&66.2&69.4&66.3\cellcolor[HTML]{EFEFEF}&60.2\cellcolor[HTML]{EFEFEF}&51.5&79.0&48.6&59.7\cellcolor[HTML]{EFEFEF}&61.4\cellcolor[HTML]{EFEFEF}\\
&+ ours&78.6\cellcolor[HTML]{EFEFEF}\textsubscript{\textcolor{red}{(+0.8)}}&71.3&66.2&40.8&90.1&67.4&71.6&67.9\cellcolor[HTML]{EFEFEF}\textsubscript{\textcolor{red}{(+1.6)}}&60.5\cellcolor[HTML]{EFEFEF}\textsubscript{\textcolor{red}{(+0.3)}}&58.1&84.1&51.0&64.4\cellcolor[HTML]{EFEFEF}\textsubscript{\textcolor{red}{(+4.7)}}&63.8\cellcolor[HTML]{EFEFEF}\textsubscript{\textcolor{red}{(+2.4)}}\\
\hline
\end{tabular}
}
}
\caption{Performance comparison of different methods across various scales (Small, Medium, Large) on multiple benchmark datasets. The \textit{Baseline} (DFN) represents the original implementation, \textit{+VLM-150M} indicates normal training on our dataset using VLM-150M, and \textit{+HQ-CLIP} represents our improved training approach on the same dataset. Subscripts $S$, $M$, and $L$ denote the subset sizes of VLM-150M used (Small, Medium, and Large respectively). Red subscripts indicate performance gains relative to the Baseline for the same scale.}

\label{tab:main_comparison_ab}
\end{table*}
\begin{table*}[htbp]
\centering
\setlength{\abovecaptionskip}{0.05cm}
\setlength{\belowcaptionskip}{0.05cm}

\resizebox{\textwidth}{!}{%
\renewcommand\arraystretch{1.1}
\setlength\tabcolsep{3.5pt}{
\begin{tabular}{ll|c|ccccccc|c|cccc|c}
\hline
\multirow{2}{*}{\begin{tabular}[c]{@{}l@{}}DataComp\\Scale\end{tabular}} & \multirow{2}{*}{Methods} & \multirow{2}{*}{IN} & \multicolumn{7}{c|}{IN dist. shifts} & VTAB & \multicolumn{4}{c|}{Retrieval} & \multirow{2}{*}{\begin{tabular}[c]{@{}c@{}}Average over \\\textbf{38}\cellcolor[HTML]{EFEFEF} datasets\end{tabular}}\\
 && {}& {V2} & A& O& R& S& ObjNet & Avg.\cellcolor[HTML]{EFEFEF}& {Avg.\cellcolor[HTML]{EFEFEF}} & {COCO} & Flickr & WinoGAViL & Avg.\cellcolor[HTML]{EFEFEF}& {}\\ \hline
\multirow{2}{*}{XLarge} & OpenAI-ViT-L &75.5\cellcolor[HTML]{EFEFEF}&69.9&70.7&32.3&87.8&59.6&69.0&64.9\cellcolor[HTML]{EFEFEF}&58.6\cellcolor[HTML]{EFEFEF}&45.7&75.1&41.4&58.6\cellcolor[HTML]{EFEFEF}&58.6\cellcolor[HTML]{EFEFEF}\\
&+Ours&76.5$_{\textcolor{red}{(+1.0)}}$\cellcolor[HTML]{EFEFEF}&70.4&70.4&36.4&88.3&61.2&68.1&65.8$_{\textcolor{red}{(+0.9)}}$\cellcolor[HTML]{EFEFEF}&60.8$_{\textcolor{red}{(+2.2)}}$\cellcolor[HTML]{EFEFEF}&56.8&85.8&56.5&66.3$_{\textcolor{red}{(+7.7)}}$\cellcolor[HTML]{EFEFEF}&63.7$_{\textcolor{red}{(+5.1)}}$\cellcolor[HTML]{EFEFEF}\\
\hline
\end{tabular}
}
}
\caption{Fine-tuning a ViT-L model on VLM-1B, our proposed \textbf{1.4 billion} pair dataset corresponding to the XLarge scale. Our method, HQ-CLIP, demonstrates notable improvements over the OpenAI baseline across a wide range of evaluation benchmarks.} 
\label{tab:main_ft}
\vspace{-3mm}
\end{table*}
\begin{table}[t!]
\setlength{\abovecaptionskip}{0.05cm}
\setlength{\belowcaptionskip}{0.05cm}
\centering
\resizebox{\linewidth}{!}{
\small
\begin{tabular}{l|cccc}
\hline
CLIP models& MMBench & MME    & MMStar & SEED \\
\hline
LAION-400M (ViT-B)    & \textbf{54.6}       & 1402.9 & 29.1   & 53.7          \\
DataComp (ViT-B)    & 50.5       & 1450.3 & 27.7   & 53.5          \\
DFN$^\dagger$ (ViT-B)    & 47.1       & 1452.1 & 28.3   & 50.6          \\
Ours (ViT-B)   & 52.8       & \textbf{1574.0} & \textbf{29.7}   & \textbf{53.8}          \\\hline
\end{tabular}
}
\caption{Performance comparison of \textbf{LLaVA1.5} using different CLIP vision encoders as the vision tower. }
\label{tab:llava}
\vspace{-3mm}
\end{table}

\noindent\textbf{Hard Negative Identification.} While negative samples are abundant in web-crawled datasets, hard negatives are crucial for CLIP's final performance. However, in conventional contrastive learning, negative samples are simply positive samples of other instances, making their difficulty uncontrollable. LVLM enables us to generate controlled hard negatives. Inspired by NegCLIP~\cite{aro}, we initially attempted to integrate these hard negative descriptions and tags by directly concatenating them into the text set. However, this naive approach demonstrated suboptimal performance.

We attribute this limitation to two key factors: the LVLM-generated hard negatives significantly outnumber positive samples, leading to dataset imbalance, and the simultaneous optimization of standard CLIP loss and hard negative identification introduces conflicting learning signals. To address this challenge, we decouple hard negative identification as an independent loss component. During each training iteration, we first compute the standard CLIP contrastive losses $\mathcal{L}_{\text{img}\rightarrow\text{txt}}$ and $\mathcal{L}_{\text{txt}\rightarrow\text{img}}$. The hard negative identification loss is subsequently computed as follows:

\vspace{-2mm}
\begin{equation}
    \label{eq:hni_loss}
    \mathcal{L}_{\text{HNI}} = -\frac{1}{N} \sum_{i=1}^N \log \frac{k_i\exp(\mathbf{x}_i^\top \mathbf{y}_i/\tau)}{\exp(\mathbf{x}_i^\top \mathbf{y}_i/\tau) + \sum_{j=1}^{N^-} \exp(\mathbf{x}_i^\top \mathbf{y}_j^-/\tau)},
\end{equation}
\vspace{-2mm}

where $\mathbf{y}_j^-$ represents embeddings of synthetic hard negative descriptions/tags, and $N^-$ indicates the number of hard negatives per instance. The gating parameter $k_i$ implements our curriculum learning strategy, defined as:

\begin{equation}
    \label{eq:k_hni_loss}
    k_i = \begin{cases} 
    1, & \text{if } i = \arg\max_{j} (\mathbf{x}_i^\top \mathbf{y}_j) \\
    0, & \text{otherwise}
    \end{cases}.
\end{equation}

This gating mechanism automatically suspends $\mathcal{L}_{\text{HNI}}$ optimization when the model fails to correctly classify standard negative samples, ensuring foundational discrimination capabilities precede hard negative learning. 

\noindent\textbf{Short Tag Classification.} While detailed textual descriptions enhance caption richness, excessive information density can obscure crucial categorical semantics. Some tasks may only require recognition of the primary object. For instance, ImageNet classification typically employs concise prompts like 'a photo of [category]' without additional details.
Inspired by ~\cite{classification}, we introduce a dual-stream learning framework that concurrently processes 1) Full descriptions for comprehensive attribute understanding and 2) Concise categorical tags for class recognition. This dual-path maintains the model's capacity for both fine-grained analysis and categorical identification, ensuring compatibility with diverse evaluation paradigms.

Our approach first analyzes the frequency distribution of semantic class tags across the entire dataset. We construct a tag vocabulary $V$ by selecting the top-$K$ most frequent tags. Considering that each image may correspond to multiple tags, we employ a multi-label binary cross-entropy loss for training an auxiliary classifier. Formally, given a tag set $\{t^+_1, t^+_2, ..., t_{N_t^+}^+\}$, we generate a multi-hot vector $\hat{{z}} \in \{0,1\}^K$ where $\hat{z}^k = 1$ indicates the presence of the $k$-th vocabulary tag. The classification loss is computed as:

\vspace{-2mm}
\begin{equation}
    \label{eq:cls_loss}
    \mathcal{L}_{\text{cls}} = -\frac{1}{N} \sum_{i=1}^N \sum_{k=1}^K \left[ \hat{z}_i^k \log \sigma(z_i^k) + (1 - \hat{z}_i^k) \log (1 - \sigma(z_i^k)) \right],
\end{equation}
\vspace{-2mm}

where $z = \mathcal{M}_{\text{cls}}(x_i)$ denotes the classifier outputs, $x_i$ represents the image embedding, $\mathcal{M}_{\text{cls}}$ is a multi-layer perceptron classifier head, and $\sigma(\cdot)$ is the sigmoid function.

\noindent\textbf{Total loss Function.} Our complete optimization objective combines the aforementioned losses:

\vspace{-2mm}
\begin{equation}
    \label{eq:total_loss}
    \mathcal{L}_{\text{total}} = 0.5\mathcal{L}_{\text{img}\rightarrow\text{txt}} + 0.5\mathcal{L}_{\text{txt}\rightarrow\text{img}} + \alpha\mathcal{L}_{\text{HNI}} + \beta\mathcal{L}_{\text{cls}}.
\end{equation}
\vspace{-2mm}

Since our dataset refinement focuses exclusively on textual enhancement without introducing new visual content, $\mathcal{L}{\text{HNI}}$ and $\mathcal{L}{\text{cls}}$ operate solely in the image-to-text direction. The combination of these objectives yields a specialized CLIP model that fully exploits the multi-grained supervision signals generated by our LVLM-driven pipeline.

\section{Experiments}
\subsection{Setup}
\noindent\textbf{Data.} Our experimental framework adopts the dataset configuration from DFN~\cite{dfn} and DataComp~\cite{datacomp}, utilizing the CommonPool corpus as the foundational data source. CommonPool aggregates web-crawled image-text pairs from Common Crawl dumps spanning 2014-2022. We offer three standardized benchmark scales: small (12.8M pairs), medium (128M pairs), and large (1.28B pairs). 

To ensure direct comparability with DFN, we employ their filtered CommonPool subset as training data. The DFN benchmark provides medium- and large-scale configurations containing 19.2M and 192M candidate pairs respectively. However, partial URLs are inaccessible, yielding effective dataset sizes of 14.7M (medium) and 146.6M (large) pairs in our implementation. We construct a small-scale baseline by random sampling 1/10 (1.47M pairs) from the DFN medium subset.
For comprehensive evaluation fairness, we report both the original DFN benchmark results as published and our reproduced outcomes (marked with $\dagger$) using acquired subsets. Details are provided in the Appendix. Ablation experiments is conducted on a medium scale.

\noindent\textbf{Training}.
We adopt the same training configurations as DFN~\cite{dfn}, including optimizer type, batch size, learning rate, weight decay, and learning rate scheduler. For large-scale experiments, we increase the number of training epochs to accommodate richer caption information, setting the total seen samples to 3.2 billion. Both our HQ-CLIP and reproduced DFN implementations maintain identical hyperparameter settings throughout all experiments. We utilize the open clip~\cite{open_clip} codebase for our implementation. 

\noindent\textbf{Evaluation}.
Our evaluation employs two benchmarks. For zero-shot classification and retrieval, we follow DataComp's protocol~\cite{datacomp}, which evaluates five key metrics: ImageNet-1K (IN), IN distribution shifts (IN-shifts), Vision Task Adaptation Benchmark (VTAB)~\cite{vtab}, Retrieval performance, and Average score across \textbf{38} diverse datasets. Additionally, we employed several multimodal benchmarks, including MME\cite{mme}, MMBench-En\cite{mmbench}, MMStar\cite{mmstar}, and SEEDBench-IMG\cite{seedbench}, using the VLMEvalKit~\cite{vlmevalkit}.
To assess fine-grained visual understanding, we utilize the ARO Benchmark~\cite{aro} with two new tasks: Visual Genome Attributions and Visual Genome Relations.

\setlength{\abovecaptionskip}{0.05cm}
\setlength{\belowcaptionskip}{0.05cm}
\begin{figure}[t!]
    \centering
    \begin{subfigure}[b]{0.45\linewidth}
        \includegraphics[width=\linewidth]{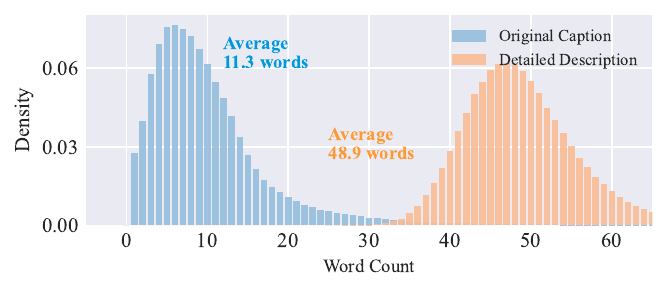}
        \caption{Length distribution of captions and detailed descriptions.}
        \label{fig:desc_len}
    \end{subfigure}
    \hfill
    \begin{subfigure}[b]{0.45\linewidth}
        \centering
        \includegraphics[width=0.8\linewidth]{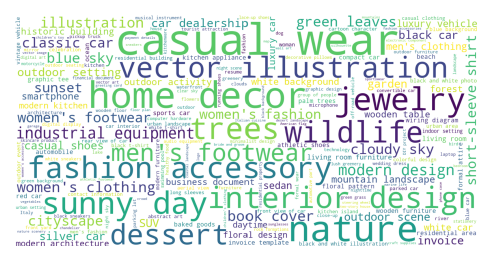}
        \caption{Word cloud of semantic class tags.}
        \label{fig:wordcloud}
    \end{subfigure}
    \caption{Analysis of dataset characteristics.}
    \label{fig:dataset_analyze}
\end{figure}
\setlength{\abovecaptionskip}{0.05cm}
\setlength{\belowcaptionskip}{0.05cm}
\begin{figure}[t!]
  \begin{minipage}[c]{0.48\linewidth}  
    \centering
    \small  
    \renewcommand\arraystretch{1.0}
    \setlength\tabcolsep{5.5pt}
    \resizebox{0.76\linewidth}{!}{
    \begin{tabular}{@{}lcc@{}}  
      \toprule
      Caption &CLIP&GPT-4o\\
      \midrule
      Origin & $25.4_{\pm\textcolor{gray}{4.3}}$ & $7.8_{\pm\textcolor{gray}{2.4}}$ \\
      w/o SFT & $27.8_{\pm\textcolor{gray}{4.4}}$ & $8.0_{\pm\textcolor{gray}{2.1}}$ \\
      w/ SFT & $\textbf{29.0}_{\pm\textcolor{gray}{4.4}}$ & $\textbf{8.4}_{\pm\textcolor{gray}{2.0}}$ \\
      Negative & $20.7_{\pm\textcolor{gray}{4.9}}$ & $0.5_{\pm\textcolor{gray}{1.1}}$ \\
      \bottomrule
    \end{tabular}
    }
    \subcaption{GPT-4o prompt: ``\textit{Given a title \{\}, rate if the title matches the image. Output a score from 0 to 10..."}}
    \vspace{-2mm}
    \label{subfig:a}
  \end{minipage}
  \hfill  
  \begin{minipage}[c]{0.48\linewidth}  
    \centering
    \includegraphics[width=32mm,height=29.1mm]{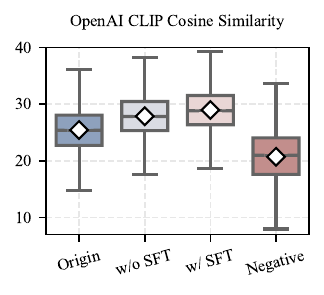} 
    \vspace{-2mm}
    \subcaption{Box plot visualization.}
    \label{subfig:b}
  \end{minipage}  
    \begin{minipage}[c]{\linewidth}  
    \centering
    \small
    \resizebox{\linewidth}{!}{
    \setlength\tabcolsep{2.5pt}
    \definecolor{mygray}{rgb}{0.95,0.95,0.95}
    \begin{tabular}{c|ccccc|>{\columncolor{mygray}}c}
    \toprule
    Caption & COCO I2T & COCO T2I & Flickr30K I2T & Flickr30K T2I & ImageNet & Average\\
    \midrule
    Origin     & 22.0 & 13.9 & 35.3 & 24.3 & 37.6 & 26.6 \\
    without SFT    & 30.1$_{(\textcolor{red}{+8.1})}$ & \textbf{19.6}$_{(\textcolor{red}{+5.7})}$ & 46.9$_{(\textcolor{red}{+11.6})}$ & 33.8$_{(\textcolor{red}{+9.5})}$ & 40.0$_{(\textcolor{red}{+2.4})}$ & 34.0$_{(\textcolor{red}{+7.4})}$ \\
    with SFT     & \textbf{32.0}$_{(\textcolor{red}{+10.0})}$ & 19.4$_{(\textcolor{red}{+5.5})}$ & \textbf{52.7}$_{(\textcolor{red}{+17.4})}$ & \textbf{36.5}$_{(\textcolor{red}{+12.2})}$ & \textbf{40.2}$_{(\textcolor{red}{+2.6})}$ & \textbf{36.2}$_{(\textcolor{red}{+9.6})}$ \\
    \bottomrule
    \end{tabular}    
    }
    \subcaption{Performance for ViT-B/32 trained on corresponding 14.7M datasets.}
    \label{subfig:c}
\label{tab:rebuttal2}
\end{minipage}
\caption{Quality analysis of original captions, LVLM-generated descriptions (with and without SFT), and negative descriptions.}
\label{fig:data_quality}
\end{figure}
\begin{table*}[htbp]
\centering
{
\small
\begin{tabular}{l|ccccc|cc}
\hline
Methods& IN& IN-Shifts & VTAB& Retrieval & Avg. over 38 datasets & Attr. & Relation \\
\hline
Baseline & 37.6& 29.7 & 37.8 & 28.6 & 36.8\cellcolor[HTML]{EFEFEF} & 54.2& 53.2 \\
+ mixed training & 40.2 & \textbf{32.7} & 41.2 & 37.3 & 39.9\cellcolor[HTML]{EFEFEF} & 59.8& 52.4 \\
+ hard negative identification & 40.1 & 32.5 & 41.6 & 38.1 & 40.7\cellcolor[HTML]{EFEFEF} & 60.0& \textbf{54.6}\\
+ short-tag classification & \textbf{40.5} & 32.3 & \textbf{42.7} & \textbf{38.4} & \textbf{41.1}\cellcolor[HTML]{EFEFEF} & \textbf{61.1}& 54.4
\\\hline
\end{tabular}
}
\caption{Ablation study of our proposed methods. The experiments are conducted on medium-scale dataset.}
\label{tab:main_ablation}
\end{table*}

\subsection{Dataset Analyze}
Fig.~\ref{fig:dataset_analyze} presents the comparative length distributions of VLM-150M's detailed descriptions and original captions. The enriched text exhibit a 4$\times$ greater average length compared to raw captions. 
Furthermore, we evaluate data quality using three metrics: a) Image-text cosine similarity with OpenAI CLIP-Large; b) GPT-4o ratings of synthetic captions, following \cite{whatif}; c) Zero-shot performance of CLIP models trained on corresponding synthetic data, following Data Filtering Networks (DFN). The data covered by evaluations a, b, and c consist of 1M, 10K, and 147M samples, respectively. Fig.\ref{subfig:a} and \ref{subfig:b} show that our method improves data quality, while Fig.\ref{subfig:c} shows that CLIP models trained on SFT-enhanced data are the best.

\subsection{Comparison with State-of-the-art}
\noindent\textbf{Datacomp benchmark evaluation}. We conduct comprehensive evaluations across 38 classification and retrieval tasks, benchmarking against state-of-the-art data filtering and re-captioning approaches. As summarized in Table~\ref{tab:main_comparison}, our method demonstrates consistent performance advantages over competitors at all scales (small/medium/large). Following standard practice where most baseline methods utilize CommonPool subsets, we adopt DFN~\cite{dfn} as our primary baseline.
Under identical hyper-parameter configurations, our method demonstrates substantial retrieval performance gains over DFN$\dagger$, achieving an improvement of +8.6\% on COCO and +11.7\% on Flickr30K at equivalent dataset scales. Remarkably, our large-scale implementation even outperforms DFN's 2 Billion data model (DFN-2B: COCO 51.9\%, Flickr 77.3\%) while operating with a significantly smaller 150 million scale dataset, achieving superior metrics of 52.5\% and 77.9\%, respectively. 

To demonstrate the scalability of our approach, we further provide results at the \textbf{XLarge} scale. We refine DFN-XLarge to develop VLM-1B, which contains 1.4 billion high-quality samples. Considering computational constraints, CLIPA~\cite{li2023clipav2scalingcliptraining} was employed for training on both, and only baseline and our full method's results are reported.

\noindent\textbf{LLaVA benchmark evaluation}. To better investigate the impact of the proposed dataset and training framework on visual understanding capabilities, we experimented using LLaVA1.5~\cite{llava1.5} and VLMEvalKit~\cite{vlmevalkit}. As shown in Table~\ref{tab:llava}, we replace the standard vision tower in LLaVA1.5 with our trained CLIP vision encoder and then repeat the pretraining and fine-tuning processes exactly as described in the original document. With a comparable training dataset scale, our models' performance surpasses other ViT-B models across multiple multimodal benchmarks.

\begin{table}[t!]
\setlength{\abovecaptionskip}{0.05cm}
\setlength{\belowcaptionskip}{0.05cm}
\centering
\small
\begin{tabular}{l|ccccc}
\hline
Ratio (\%)                                                              & 0    & 25   & 50   & 75   & 100  \\
\hline
ImageNet                                                           & 37.6 & 40.0 & \textbf{40.9} & 40.2 & 35.3 \\
ImageNet-Shifts                                                    & 29.7 & 31.9 & \textbf{32.8} & 32.7 & 29.9 \\
VTAB                                                               & 37.8 & 39.7 & 39.7 & \textbf{41.2} & 38.2 \\
Retrieval                                                          & 28.6 & 34.3 & 35.7 & \textbf{37.3} & 33.7 \\
\hline
Avg. over 38 datasets & 36.8 & 38.8 & 39.8 & \textbf{39.9} & 36.9\\
\hline
\end{tabular}
\caption{Ablation study on raw/enriched text mixing ratios}
\vspace{-3mm}
\label{tab:ablation_mixed_ratio}
\end{table}

\begin{figure}[htbp]
    \centering
    \includegraphics[width=1.0\linewidth]{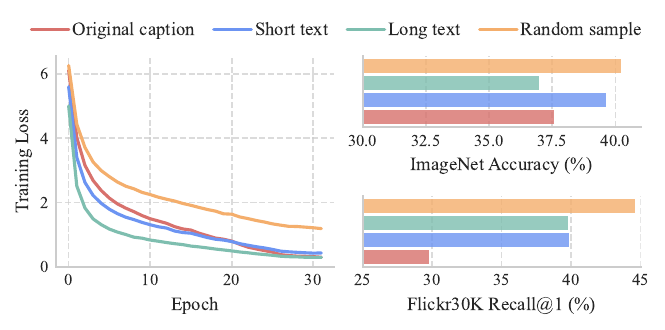}
    \caption{The loss curve and performance of using generated \textbf{long text}, generated \textbf{short text}, and \textbf{randomly sampled} short text from generated long text. Interestingly, the method with the highest loss achieves the highest ImageNet classification performance, while the method with the lowest loss performs the worst. Notably, the Flickr30K retrieval accuracy of the random sample is significantly higher than that of the others.}
    \label{fig:select}
\end{figure}

\subsection{Ablation Study}
\label{sec:ablation}

\noindent\textbf{Ablation on Main Components}. 
Table~\ref{tab:main_comparison_ab} quantifies the performance improvements attributable to both the VLM-150M dataset and the proposed HQ-CLIP framework. 
The VLM-150M dataset consistently enhances CLIP performance across all scales, demonstrating its superior data quality compared to DFN. 
Furthermore, HQ-CLIP delivers additional gains of \textbf{+0.7/+1.2/+2.0} on Small/Medium/Large scales when applied to VLM-150M. 
This cross-scale consistency validates HQ-CLIP's efficacy, while the progressive performance increase suggests that larger models benefit more extensively from its multi-grained supervision signals (negative descriptions and short tags).

\noindent\textbf{Ablation on HQ-CLIP}. As detailed in Table~\ref{tab:main_ablation}, we conduct systematic ablation studies at a medium scale using our reproduced DFN as the baseline. The initial integration of our VLM-150M dataset through mixed training demonstrates fundamental effectiveness, yielding a significant performance enhancement of \textit{3.1}\%. Furthermore, the introduction of hard-negative descriptions and the short-tag classification paradigm substantially improves performance, resulting in an additional enhancement of \textit{1.2}\%.

\noindent\textbf{Ablation on Text Length.} Excessive text length can impair CLIP training effectiveness \cite{longclip}. We investigate optimal text comparing three strategies: full-length generated descriptions, generated short texts, and randomly sampled short texts from long descriptions (with original captions as baseline). As shown in Fig.\ref{fig:select}, the method with the highest loss achieves the best zero-shot generalization, while the method with the lowest loss performs the worst. This is particularly notable in Flickr30K retrieval, where random sampling significantly outperforms other approaches. This suggests that over-detailed descriptions reduce contrastive learning difficulty. Given that phenomenon and CLIP's inherent 77-token limitation \cite{clip}, we adopt random short-text sampling from generated descriptions. The punctuation used to segment sentences is generated by LVLM.


\noindent\textbf{Ablation study on mix ratios and weights}. We conduct an ablation analysis examining the mixed training ratio $r$, hard-negative identification loss weight $\alpha$, and short-tag classification loss weight $\beta$. As shown in Tables~\ref{tab:ablation_mixed_ratio}, We empirically identify 75\% as the optimal mixing ratio. Experiments about $\alpha$ and $\beta$ are provided in the supplementary materials.

\section{Limitations and conclusion}
We present an efficient LVLM-driven dataset refinement pipeline that transforms DFN-Large into \textbf{VLM-150M} - a high-quality image-text dataset featuring multi-grained descriptions. These complementary captions enable our proposed training paradigm, \textbf{HQ-CLIP}, which extends conventional contrastive learning through negative descriptions and short-tag supervision. Comprehensive evaluations demonstrate HQ-CLIP's superior performance across zero-shot classification, retrieval, and understanding tasks. When substituted as LLaVA's vision encoder, HQ-CLIP outperforms CLIP models of comparable pre-training scale, highlighting its potential for advancing LVLM development.

While HQ-CLIP achieves SoTA performance at comparable training scales, our VLM-150M-based solution still lags behind the capabilities of DFN-5B. Continued efforts to scale VLM-150M to billions of samples and upgrade HQ-CLIP to ViT-L architectures remain imperative. We hope that future works will investigate optimal training strategies for CLIP models by leveraging multi-grained bidirectional descriptions, as well as methodologies for advancing LVLM performance through VLM-150M integration. We anticipate that this work will serve as a foundational resource for advancing multimodal learning.
\clearpage
{
    \small
    \bibliographystyle{ieeenat_fullname}
    \bibliography{main}
}

\maketitlesupplementary
\section{Experiments}
\begin{table}[htbp]
\centering
\resizebox{\linewidth}{!}{
\setlength\tabcolsep{2.5pt}
\small
\begin{tabular}{l|ccc}
\hline
DataComp Scale      & Small    & Medium   & Large    \\\hline
CommonPool size     & 12.8M    & 128M     & 1.28B    \\
Original DFN size   & -        & 19.2M    & 192M     \\
Reproduced DFN size & 1.47M    & 14.7M    & 147M     \\
Model               & ViT-B/32 & ViT-B/32 & ViT-B/16 \\
Batch size          & 4096     & 4096     & 8192      \\\hline
\end{tabular}
}
\caption{Training setup and dataset scale.}
\vspace{-3mm}
\label{tab:setup}
\end{table}
\begin{table}[t!]
\renewcommand\arraystretch{1.1}
\setlength{\abovecaptionskip}{0.05cm}
\setlength{\belowcaptionskip}{0.05cm}
\centering
\resizebox{0.65\linewidth}{!}{%
\setlength\tabcolsep{3.5pt}{
{
\small
\begin{tabular}{l|l|cc}
\hline
Scale                   & Methods & Attribution & Relation  \\
\hline
\multirow{2}{*}{Medium} & DFN$^\dagger$     & 54.2        & 53.2      \\
                        & Ours    & \textbf{61.1}        & \textbf{54.4}     \\
\hline
\multirow{2}{*}{Large}  & DFN$^\dagger$     & 55.1        & 47.2     \\
                        & Ours    & \textbf{65.1 }       & \textbf{61.3}  
\\\hline
\end{tabular}
}
}
}
\caption{Comparison of attribution and relation metrics in the ARO benchmark~\cite{aro}.}
\label{tab:main_aro}
\end{table}
\begin{table}[t!]
\centering
\small
\begin{tabular}{l|cccc}
\hline
Number of classes                                                  & 3000 & 10000 & 30000 & 90000 \\
\hline
ImageNet                                                           & 39.2 & \textbf{40.8} &40.5         & 40.6  \\
ImageNet-Shifts                                                    & 31.1 & \textbf{32.9} &32.8         & 32.8 \\
VTAB                                                               & 40.3 & 40.5          &40.5         & \textbf{42.3} \\
Retrieval                                                          & 36.6 & 37.4          &\textbf{37.7}& 37.3 \\
\hline
\begin{tabular}[c]{@{}l@{}}Average over\\ 38 datasets\end{tabular} & 39.9 & 40.2          &40.1         & \textbf{40.5} \\
\hline
\end{tabular}
\caption{Ablation study on the number of classes.}
\label{tab:ablation_num_classes}
\end{table}
\begin{table}[t!]
\centering
\small
\begin{tabular}{l|cccc}
\hline
$\beta$                                                                & 1   & 10  & 100 &1000 \\ \hline
ImageNet                                                           &40.1 &40.0 &\textbf{40.7} &40.6 \\
ImageNet-Shifts                                                    &32.4 &32.4 &\textbf{33.2} &32.1 \\
VTAB                                                               &44.1 &44.3 &\textbf{44.7} &44.1 \\
Retrieval                                                          &37.2 &36.9 &\textbf{37.5} &36.8 \\ \hline
\begin{tabular}[c]{@{}l@{}}Average over\\ 38 datasets\end{tabular} &40.1 &39.9 &\textbf{40.5} &40.2 \\ \hline
\end{tabular}
\caption{Ablation study on the weight of $\mathcal{L}_{STC}$.}
\label{tab:ablation_class_weight}
\end{table}
\begin{table}[htbp]
\centering
\begin{tabular}{l|cccc}
\hline
$\alpha$                                                                & 0.1 & 0.2 & 0.5 & 1  \\ \hline
ImageNet                                                           &40.6 &\textbf{40.7} &40.1 &40.2\\
ImageNet-Shifts                                                    &\textbf{32.7} &32.5 &32.5 &32.2\\
VTAB                                                               &40.8 &41.2 &\textbf{41.6} &41.3\\
Retrieval                                                          &38.7 &37.7 &\textbf{38.1} &\textbf{38.1}\\ \hline
\begin{tabular}[c]{@{}l@{}}Average over\\ 38 datasets\end{tabular} &40.1 &39.9 &\textbf{40.7} &40.0\\ \hline
\end{tabular}
\caption{Ablation study on the weight of $\mathcal{L}_{HNI}$.}
\label{tab:ablation_hni_k}
\end{table}
\begin{table*}[htbp]
\centering
\resizebox{\textwidth}{!}{%
\renewcommand\arraystretch{1.1}
\setlength\tabcolsep{3.5pt}{
\begin{tabular}{l|c|cccc|cccccccccc}
\hline
       & \begin{tabular}[c]{@{}c@{}}Dataset\\ size\end{tabular} & IN            & INv2          & COCO          & Flickr        & Caltech101    & CIFAR100      & SVHN          & DTD           & OxPet         & Flowers102    & EuroSAT       & RESISC45      & Camelyon      & Average\cellcolor[HTML]{EFEFEF}       \\\hline
       
VeCLIP~\cite{veclip} & 200M                                                   & 64.6          & 57.7          & \textbf{57.8} & \textbf{83.7} & 83.1          & 68.1          & 44.9          & \textbf{62.0} & 72.6          & 68.5          & 47.4          & 55.1          & \textbf{62.6} & 62.7\cellcolor[HTML]{EFEFEF}          \\
Ours   & 148M                                                   & \textbf{70.6} & \textbf{63.1} & 52.2          & 77.9          & \textbf{93.1} & \textbf{81.0} & \textbf{45.9} & 51.5          & \textbf{89.5} & \textbf{69.0} & \textbf{47.6} & \textbf{60.6} & 46.2          & \textbf{64.9}\cellcolor[HTML]{EFEFEF} \\\hline
\end{tabular}
}}
\caption{Comparison of Our Method with VeCLIP. The metrics for VeCLIP are sourced from the original paper. Our method demonstrates superior average performance.}
\label{tab:veclip}
\end{table*}
\begin{table}[htbp]
\centering
\resizebox{\linewidth}{!}{%
\renewcommand\arraystretch{1.1}
\setlength\tabcolsep{3.5pt}{
\begin{tabular}{l|c|cccc|c}
\hline
           & IN            & IN-Shifts     & VTAB          & Retrieval     & Average over 38 datasets \\\hline
VeCLIP$^*$ & 52.5          & 45.9          & 46.8          & 55.2          & 48.2\cellcolor[HTML]{EFEFEF}                     \\
Ours       & \textbf{70.6} & \textbf{57.2} & \textbf{57.6} & \textbf{60.9} & \textbf{58.6}\cellcolor[HTML]{EFEFEF}           \\\hline
\end{tabular}
}
}
\caption{Comparison of Performance on the DataComp~\cite{datacomp} Benchmark with VeCLIP. The metrics for VeCLIP were obtained by using the weights provided in its official GitHub repository, trained on the 100 Million dataset, and evaluated using the DataComp benchmark code and Hugging Face tools.}
\label{tab:veclip2}
\end{table}
\subsection{Setup}
Our experimental setup primarily follows the configuration established in DFN~\cite{dfn}. 
The original DFN methodology processes CommonPool datasets (12.8M/128M/1.28B) to derive filtered subsets of 1.92M (small), 19.2M (medium), and 192M (large) image-text pairs. 
Due to partial URL inaccessibility, we obtained reduced subsets of \textbf{1.47M (small)}, \textbf{14.7M (medium)}, and \textbf{147M (large)} pairs for our implementation. 
In model training, we strictly adhere to DFN's architectural specifications and batch size configurations. 
Notably, for the XLarge-scale model training, we employed CLIPA~\cite{li2023clipav2scalingcliptraining} to optimize computational efficiency and accelerate training convergence.

\subsection{Ablation study}

\begin{table}[htbp]
\centering
\begin{tabular}{l|ccccc}
\hline
$N^-$                                                              & 1    & 2    &3&4  \\
\hline
ImageNet                                                           & 39.9 & 39.8&\textbf{40.0}&39.5  \\
ImageNet-Shifts                                                    & \textbf{32.5} & \textbf{32.5} &32.1&32.3\\
VTAB                                                               & 41.8 & \textbf{42.0} &40.9&41.1\\
Retrieval                                                          & \textbf{38.1} & 38.0 & 37.7&37.9\\
\hline
\begin{tabular}[c]{@{}l@{}}Average over\\ 38 datasets\end{tabular} & 40.1 & \textbf{40.2} &40.1 &39.9\\
\hline
\end{tabular}
\caption{Ablation study on number of hard negative samples $M$.}
\label{tab:ablation_hni_m}
\end{table}

\noindent\textbf{Ablation study on hard-negative sample quantity.}
We investigate the optimal number of hard-negative variants per image for identification tasks. As Table~\ref{tab:ablation_hni_m} demonstrates, empirical evidence suggests the single-sample configuration emerges as optimal. Although increasing the number of samples initially appears to benefit performance metrics, practical constraints such as prohibitive GPU memory demands and computational overhead prevent further scaling. Consequently, we select one hard-negative instance as the computationally efficient yet effective solution.

\noindent\textbf{Ablation on the number of classes}. Our framework employs a frequency-based selection of the top-K most prevalent tags from the VLM-generated tag repository. As empirically validated in Table~\ref{tab:ablation_num_classes}, we systematically determine the optimal class quantity parameter $K$.

\noindent\textbf{Ablation study on loss hyperparameters $\alpha$ and $\beta$}.  
Performance sensitivity to the hard-negative identification loss weight ($\alpha$) and short tag classification loss weight ($\beta$) is quantified in Tables~\ref{tab:ablation_hni_k} and \ref{tab:ablation_class_weight}. The optimal configuration is observed at $\alpha = 0.5$ and $\beta = 10$, where both loss components contribute maximally to model effectiveness.

\subsection{Comparison with state-of-the-art method}
\noindent\textbf{ARO benchmark evaluation}. 
As shown on Tab.~\ref{tab:main_aro}, our approach exhibits superior comprehension of attribution and relation compared to the DFN$^\dagger$ baseline. By benefiting from descriptions with enhanced semantic richness and the specialized hard-negative identification loss during training, our method achieves significant and scalable performance improvements on Visual Genome attribution metrics.

\noindent\textbf{Comparison with VeCLIP}.
Given the exceptional performance claims of VeCLIP\cite{veclip} in its original publication, comprehensive benchmarking becomes imperative. However, since VeCLIP did not include DataComp benchmark results in their work, a direct comparison in our main results table (Table 2) proves infeasible. We therefore provide supplementary comparisons with more performance metrics in the Supplementary Materials between our method and the ViT-B variant of VeCLIP trained on 200 million samples (as reported in their paper), where our approach demonstrates superior comprehensive performance (Table\ref{tab:veclip}).

To facilitate rigorous benchmarking, we sought to evaluate VeCLIP under the DataComp\cite{datacomp} framework. While the authors provide clear instructions for loading their ViT-H weights, documentation gaps were identified regarding ViT-B weight implementation. Technical challenges emerged from (1) framework-specific implementation details in TensorFlow and (2) compatibility constraints with VeCLIP's text encoder architecture in the DataComp library. To address these methodological challenges, we re-implemented a PyTorch version of VeCLIP's data pipeline and modified the DataComp evaluation code.

Due to technical limitations in loading VeCLIP model weights trained on the 200M subset, our analysis employs the 100M variant for standardized DataComp benchmark comparisons (Table \ref{tab:veclip2}). HQ-CLIP significantly outperforms VeCLIP. We are actively seeking verification through direct communication with the authors' team to ensure correct comparison and sincerely welcome their insights.

\begin{figure}
    \includegraphics[width=\linewidth]{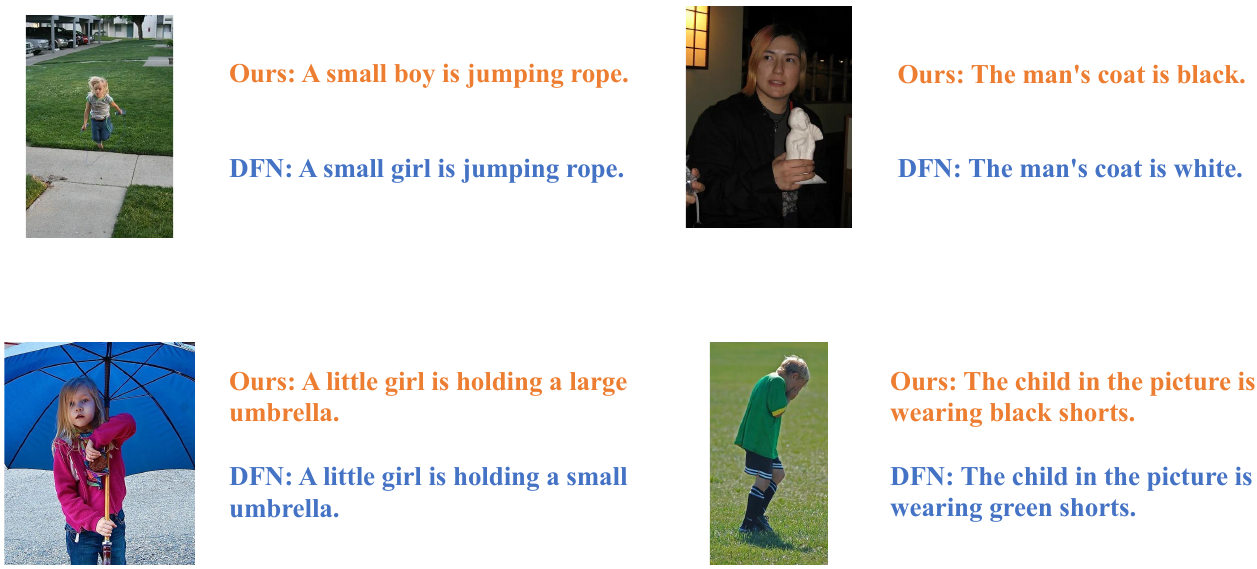}
    \caption{Comparison of recognition results between our model and DFN.}
    \label{fig:retrieval}
\end{figure}

\subsection{Recognition Results}
Figure~\ref{fig:retrieval} shows the classification results of our model compared to the DFN model. For each image, binary classification is performed using manually crafted text to demonstrate the fine-grained understanding capability of the models. Our model shows better recognition of detailed semantics in the images.

\subsection{Details of other experiments}
We showcase the full 38 dataset result for some experiments on main paper, as shown in Tab.~\ref{tab:investigation_vlms} and ~\ref{tab:main_comparison}.
\begin{table}[htbp]
\centering
\resizebox{\linewidth}{!}{
\begin{tabular}{lcccccc}
\toprule
Model & XCom2 & LLaVA & Qwen2-VL & Qwen2-VL & Qwen2-VL & Qwen2-VL \\
\midrule
Parameters & 7B & 7B & 7B & 2B & 72B & 7B \\
GPT4o SFT & \checkmark & \checkmark & \checkmark & \checkmark &  & \checkmark \\
Caption Input & \checkmark & \checkmark &  & \checkmark & \checkmark & \checkmark \\
ImageNet 1k & 41.1 & 39.9 & 37.6 & 40.8 & \textbf{41.2} & 40.2 \\
ImageNet Sketch & 30.9 & 31.1 & 26.9 & 31.9 & \textbf{31.9} & 31.7 \\
ImageNet V2 & \textbf{34.1} & 33.3 & 30.6 & 33.8 & 34.1 & 33.4 \\
ImageNet-A & 7.1 & \textbf{7.5} & 6.2 & 6.8 & 7.2 & 7.2 \\
ImageNet-O & \textbf{48.9} & 47.8 & 46.0 & \textbf{48.9} & 48.1 & 48.0 \\
ImageNet-R & \textbf{47.6} & 47.5 & 42.5 & 47.4 & 47.5 & 47.5 \\
Caltech-101 & 81.7 & 80.4 & 78.9 & 80.8 & 80.7 & \textbf{83.8} \\
CIFAR-10 & 89.8 & 88.2 & 83.8 & 88.1 & 88.4 & \textbf{89.8} \\
CIFAR-100 & 63.8 & 63.6 & 59.2 & 65.2 & 65.0 & \textbf{65.5} \\
CLEVR Counts & 14.9 & \textbf{26.2} & 13.1 & 24.3 & 17.1 & 25.0 \\
CLEVR Distance & \textbf{21.2} & 18.6 & 16.4 & 15.9 & 15.9 & 15.8 \\
SVHN & \textbf{26.8} & 10.6 & 20.4 & 21.9 & 9.8 & 23.1 \\
DTD & 28.0 & 26.1 & 22.0 & 27.7 & 27.8 & \textbf{28.7} \\
EuroSAT & 35.9 & \textbf{40.9} & 22.5 & 31.4 & 36.5 & 32.6 \\
KITTI distance & 20.5 & 28.7 & 16.7 & 27.1 & \textbf{34.2} & 32.1 \\
Oxford Flowers-102 & 38.8 & 35.8 & 39.3 & 39.3 & \textbf{39.7} & 36.3 \\
Oxford-IIIT Pet & 59.5 & 60.0 & 57.0 & 58.8 & \textbf{61.3} & 55.4 \\
PatchCamelyon & 57.5 & 54.7 & 56.8 & 52.3 & \textbf{58.7} & 53.1 \\
RESISC45 & 31.0 & 34.8 & 28.7 & \textbf{36.7} & 33.9 & 34.5 \\
FGVC Aircraft & 3.3 & 3.2 & 3.3 & 2.6 & \textbf{3.5} & 3.4 \\
Food-101 & 56.1 & 54.5 & 52.8 & \textbf{56.5} & 55.4 & 56.1 \\
GTSRB & 15.5 & 18.6 & 13.9 & 17.1 & 17.1 & \textbf{19.7} \\
MNIST & 29.8 & 22.8 & 23.4 & 29.5 & 26.1 & \textbf{31.8} \\
ObjectNet & 28.5 & \textbf{28.7} & 24.3 & 28.6 & 28.0 & 28.4 \\
Pacal VOC 2007 & 63.8 & 70.2 & 54.7 & 67.6 & 69.1 & \textbf{71.0} \\
Rendered SST2 & 50.2 & 50.1 & \textbf{50.4} & 49.9 & 49.2 & 49.7 \\
Stanford Cars & 45.3 & 44.1 & \textbf{48.9} & 45.7 & 48.2 & 42.5 \\
STL-10 & 89.9 & 89.9 & 87.1 & 89.8 & 90.0 & \textbf{90.2} \\
SUN-397 & 48.7 & 47.4 & 44.5 & 48.8 & 48.7 & \textbf{49.7} \\
Country211 & 5.0 & 4.8 & 4.5 & 5.3 & \textbf{5.3} & 5.3 \\
iWildCam & 2.9 & 2.2 & 2.3 & 2.5 & \textbf{3.5} & 2.6 \\
Camelyon17 & 57.0 & 65.8 & \textbf{67.8} & 53.1 & 66.0 & 55.8 \\
FMoW & \textbf{0.0} & \textbf{0.0} & \textbf{0.0} & \textbf{0.0} & \textbf{0.0} & \textbf{0.0} \\
Dollar Street & \textbf{49.3} & 46.1 & 47.1 & 48.2 & 48.7 & 47.4 \\
GeoDE & 73.0 & 70.5 & 66.2 & \textbf{74.0} & \textbf{74.0} & 68.8 \\
Flickr30k & 40.8 & 42.3 & 29.5 & 42.0 & 39.6 & \textbf{44.6} \\
MSCOCO & 25.3 & 18.0 & 17.2 & 26.2 & 24.2 & \textbf{26.7} \\
WinoGAViL & 43.1 & 37.7 & 36.9 & 41.6 & \textbf{46.4} & 40.5 \\
Avg. over 38 datasets & 39.6 & 39.3 & 36.3 & 39.7 & \textbf{40.1} & 39.9 \\
\bottomrule
\end{tabular}
}
\caption{Comparison of the performance of different data refinement pipelines. Compared to other LVLMs, Qwen2VL demonstrates superior performance. Despite a tenfold difference in parameter size, Qwen2VL-7B with GPT-4o SFT still exhibits performance comparable to the 72B model. Additionally, the inclusion of captions significantly enhances dataset quality.}
\label{tab:investigation_vlms_details}
\end{table}

\begin{table}[htbp]
\centering

{%
\renewcommand\arraystretch{1.1}
\setlength\tabcolsep{3.5pt}{
\begin{tabular}{lcc}
\toprule
Method & Ours & DFN \\
\midrule
DataComp scale & Large & Large \\
Dataset size & 146.6M & 146.6M \\
ImageNet 1k & \textbf{70.6} & 68.7 \\
ImageNet Sketch & \textbf{57.3} & 54.9 \\
ImageNet V2 & \textbf{63.1} & 60.0 \\
ImageNet-A & \textbf{39.1} & 29.9 \\
ImageNet-O & 43.0 & \textbf{53.5} \\
ImageNet-R & \textbf{80.1} & 75.4 \\
Caltech-101 & \textbf{93.1} & 91.2 \\
CIFAR-10 & \textbf{96.2} & 94.8 \\
CIFAR-100 & \textbf{81.0} & 79.1 \\
CLEVR Counts & \textbf{27.5} & 14.7 \\
CLEVR Distance & \textbf{22.2} & 20.0 \\
SVHN & 45.9 & \textbf{48.5} \\
DTD & \textbf{51.5} & 46.9 \\
EuroSAT & 47.6 & \textbf{49.9} \\
KITTI distance & \textbf{43.0} & 24.9 \\
Oxford Flowers-102 & 69.0 & \textbf{71.0} \\
Oxford-IIIT Pet & \textbf{89.5} & 88.7 \\
PatchCamelyon & 47.5 & \textbf{51.0} \\
RESISC45 & \textbf{60.6} & 56.0 \\
FGVC Aircraft & 11.3 & \textbf{13.2} \\
Food-101 & \textbf{87.8} & 86.2 \\
GTSRB & \textbf{54.4} & 44.2 \\
MNIST & \textbf{77.7} & 61.5 \\
ObjectNet & \textbf{60.6} & 55.0 \\
Pacal VOC 2007 & \textbf{78.8} & 75.0 \\
Rendered SST2 & \textbf{51.7} & 51.2 \\
Stanford Cars & \textbf{85.3} & 85.1 \\
STL-10 & \textbf{98.1} & 96.0 \\
SUN-397 & \textbf{69.7} & 67.2 \\
Country211 & \textbf{15.9} & 13.5 \\
iWildCam & \textbf{12.2} & 10.0 \\
Camelyon17 & 46.2 & \textbf{63.1} \\
FMoW & \textbf{15.1} & 10.9 \\
Dollar Street & \textbf{61.3} & 60.3 \\
GeoDE & \textbf{88.7} & 87.3 \\
Flickr30k & \textbf{77.9} & 68.2 \\
MSCOCO & \textbf{52.2} & 43.7 \\
WinoGAViL & \textbf{52.8} & 51.8 \\
Avg. over 38 datasets & \textbf{58.6} & 55.9 \\
\bottomrule
\end{tabular}
}
}
\caption{Training on VLM-150M yields state-of-the-art CLIP models. We evaluate these models using the DataComp evaluation protocol. For detailed comparisons on specific datasets, we also provide the reproduced results for DFN. The symbol $\dagger$ indicates the results that we reproduced. Due to some broken links in the dataset, the amount of data used in our reproduction is slightly lower than that in the original paper.}
\label{tab:main_comparison_full}
\end{table}

\section{VLM-150M}
\subsection{Examples}
\begin{figure*}[htbp]
    \centering
    \begin{subfigure}[b]{\linewidth}
        \includegraphics[width=\linewidth]{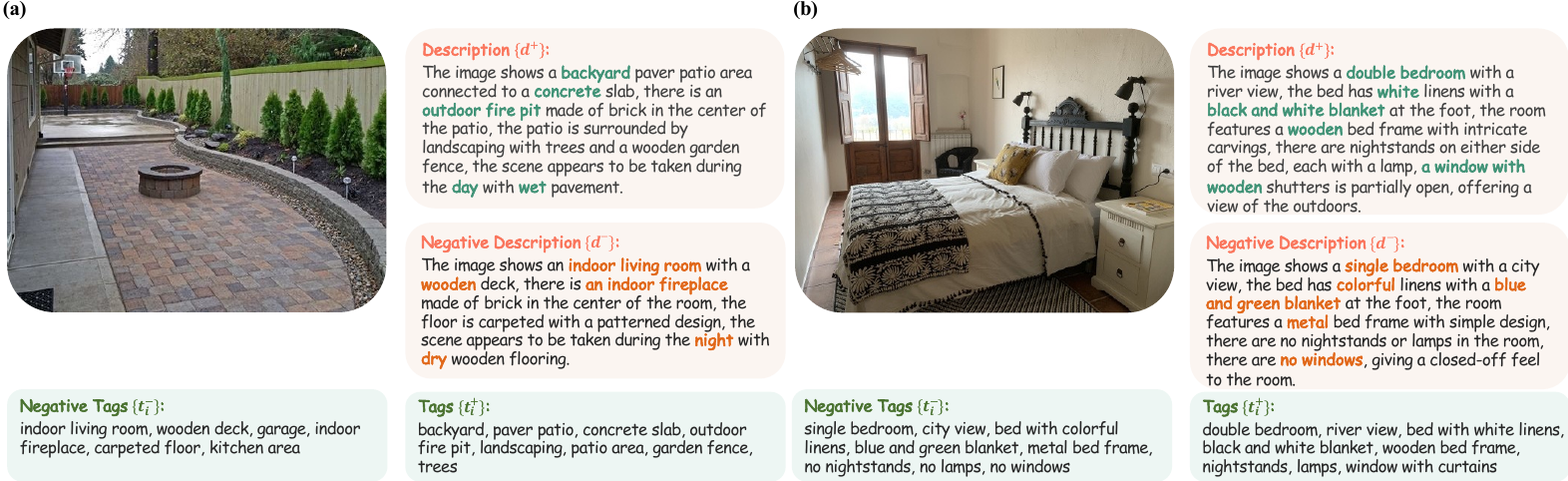}        
    \end{subfigure}
    \hfill
    \begin{subfigure}[b]{\linewidth}
        \centering
        \includegraphics[width=\linewidth]{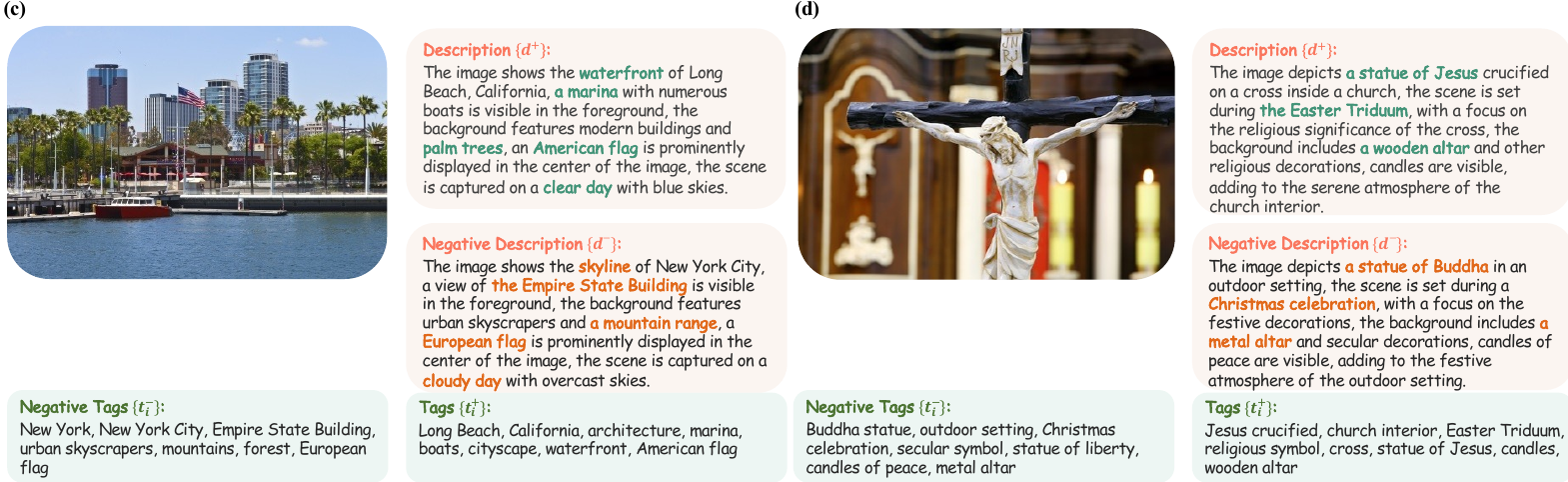}        
    \end{subfigure}
    \begin{subfigure}[b]{\linewidth}
        \centering
        \includegraphics[width=\linewidth]{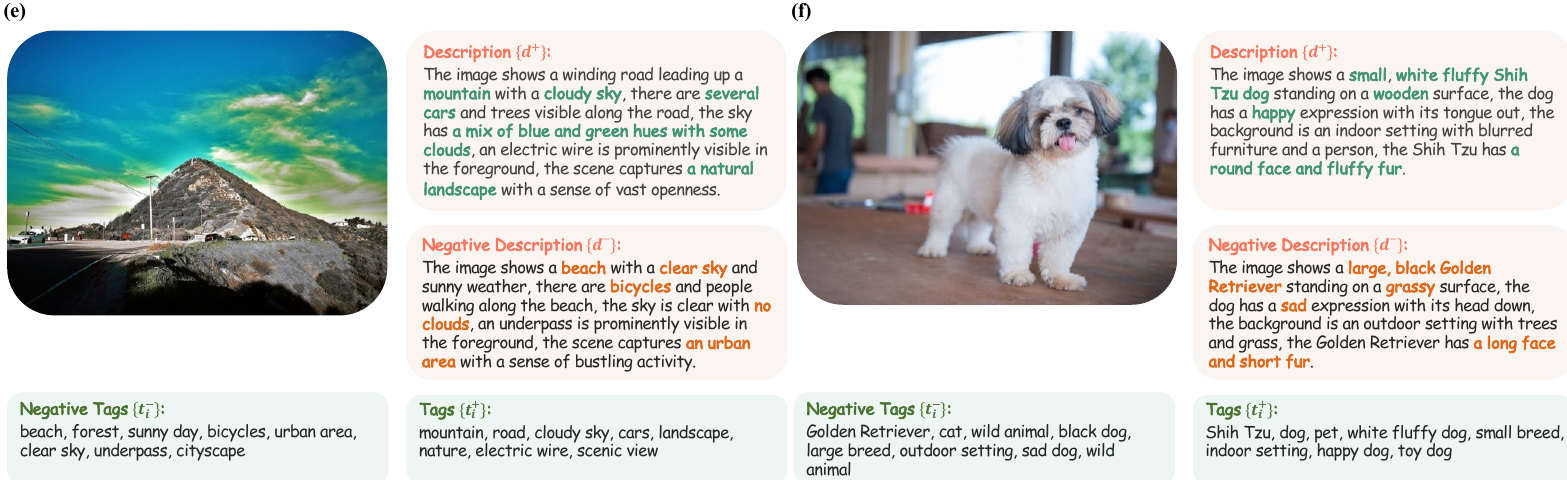}        
    \end{subfigure}
    \begin{subfigure}[b]{\linewidth}
        \centering
        \includegraphics[width=\linewidth]{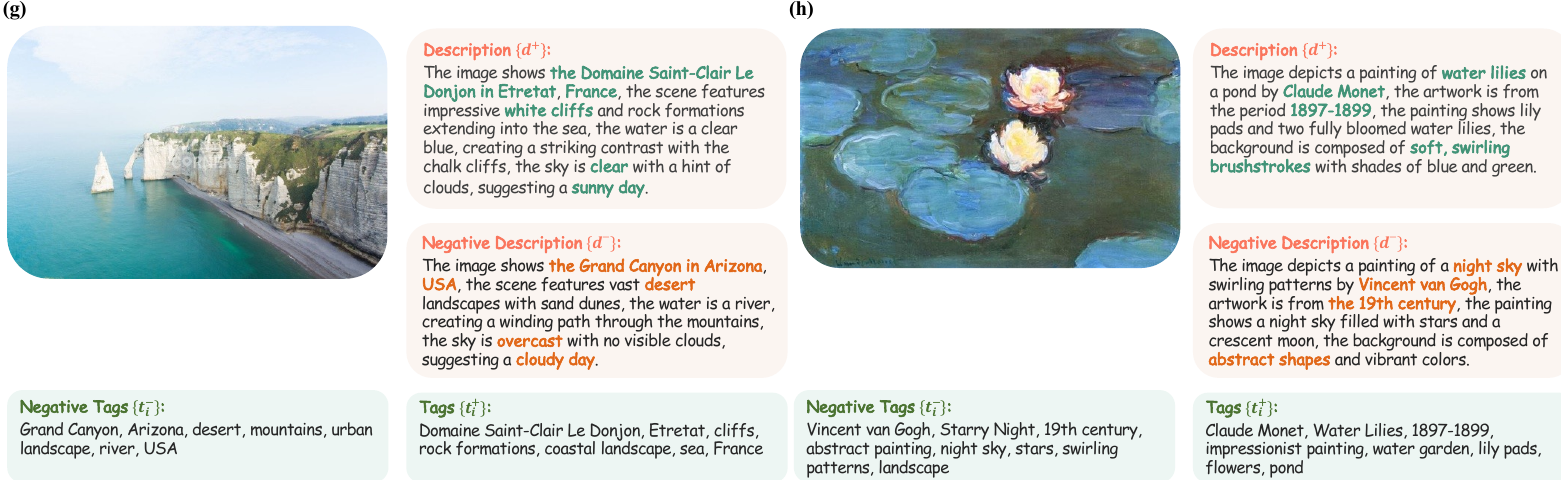}        
    \end{subfigure}
    \caption{Examples of VLM-150M.}
    \label{fig:examples}
\end{figure*}

We present some examples from the acquired dataset. As shown in Figure \ref{fig:examples}, we obtained more comprehensive annotations.

\end{document}